\documentclass[10pt,twocolumn,letterpaper]{article}

\usepackage[pagenumbers]{cvpr} 

\usepackage{times}  
\usepackage{helvet}  
\usepackage{courier}  
\usepackage[hyphens]{url}  
\usepackage{graphicx} 
\usepackage{multirow} 
\usepackage{multicol}
\usepackage{colortbl} 
\usepackage{rotating}
\usepackage{caption}
\usepackage{booktabs}
\usepackage{amsfonts}       
\usepackage{nicefrac}       
\usepackage{amsmath}
\usepackage{tabularx}
\usepackage{tcolorbox}

\usepackage{algorithm}
\usepackage{algorithmicx}
\usepackage{pifont}
\usepackage{amssymb}
\usepackage{tablefootnote}
\usepackage{float}
\usepackage{newfloat}
\usepackage{listings}
\usepackage{makecell}
 \usepackage[misc]{ifsym}
 \usepackage{cutwin}
\definecolor{cvprblue}{rgb}{0.21,0.49,0.74}
\usepackage[utf8]{inputenc} 
\usepackage[T1]{fontenc}    
\usepackage{amssymb}
\usepackage{fontawesome}
\DeclareUnicodeCharacter{2713}{\checkmark}  
\DeclareUnicodeCharacter{2717}{\ding{55}}   
\usepackage[pagebackref,breaklinks,colorlinks,allcolors=cvprblue]{hyperref}
\title{
VisuRiddles: Fine-grained Perception is a Primary Bottleneck for Multimodal Large Language Models in Abstract Visual Reasoning
}

\author{Hao Yan\textsuperscript{1} \quad Xingchen Liu\textsuperscript{1} \quad Hao Wang\textsuperscript{2} \quad Zhenbiao Cao\textsuperscript{1} \quad Handong Zheng\textsuperscript{1} \quad Liang Yin\textsuperscript{1}\\ 
Xinxing Su\textsuperscript{2} \quad Zihao Chen\textsuperscript{2} \quad Jihao Wu\textsuperscript{2} \quad Minghui Liao\textsuperscript{2}$^*$\quad \quad Chao Weng\textsuperscript{2}\\
\quad Wei Chen\textsuperscript{1}\quad Yuliang Liu\textsuperscript{1} \quad Xiang Bai\textsuperscript{1} \\
\textsuperscript{1}{Huazhong University of Science and Technology} \quad
\textsuperscript{2}{Huawei Inc.}\\
}

\begin{document}
\maketitle
\begin{abstract}
Recent strides in multimodal large language models (MLLMs) have demonstrated significant progress in many reasoning tasks, but they still fail in Abstract Visual Reasoning (AVR) tasks. Our experimental findings indicate that the core bottleneck lies not only in the reasoning capabilities of MLLMs but more critically in their absence of fine-grained perception. To address this issue, we present VisuRiddles, a dedicated resource for AVR research. It consists of (i) a benchmark, collected from real-world data, for the systematic evaluation of MLLMs' AVR capabilities, and (ii) a synthesizer, which automatically generates AVR instances enriched with perceptual descriptions and reasoning chains, enabling supervised training and deeper investigation. Building on VisuRiddles, we propose a two-stage training paradigm that progressively enhances perceptual ability and strengthens reasoning, producing the Perception-Augmented Visual Reasoner (PAVR). Experiments demonstrate that PAVR unifies perception and reasoning, substantially outperforming both open-source and commercial MLLMs, thereby underscoring fine-grained perception as the primary bottleneck in AVR. Our code and dataset will be released at \href{https://github.com/yh-hust/VisuRiddles}{https://github.com/yh-hust/VisuRiddles}
\end{abstract}

\let\thefootnote\relax\footnotetext{$^*$ Project Lead}

\begin{figure*}[t!]
  \centering
  \includegraphics[width=\linewidth]{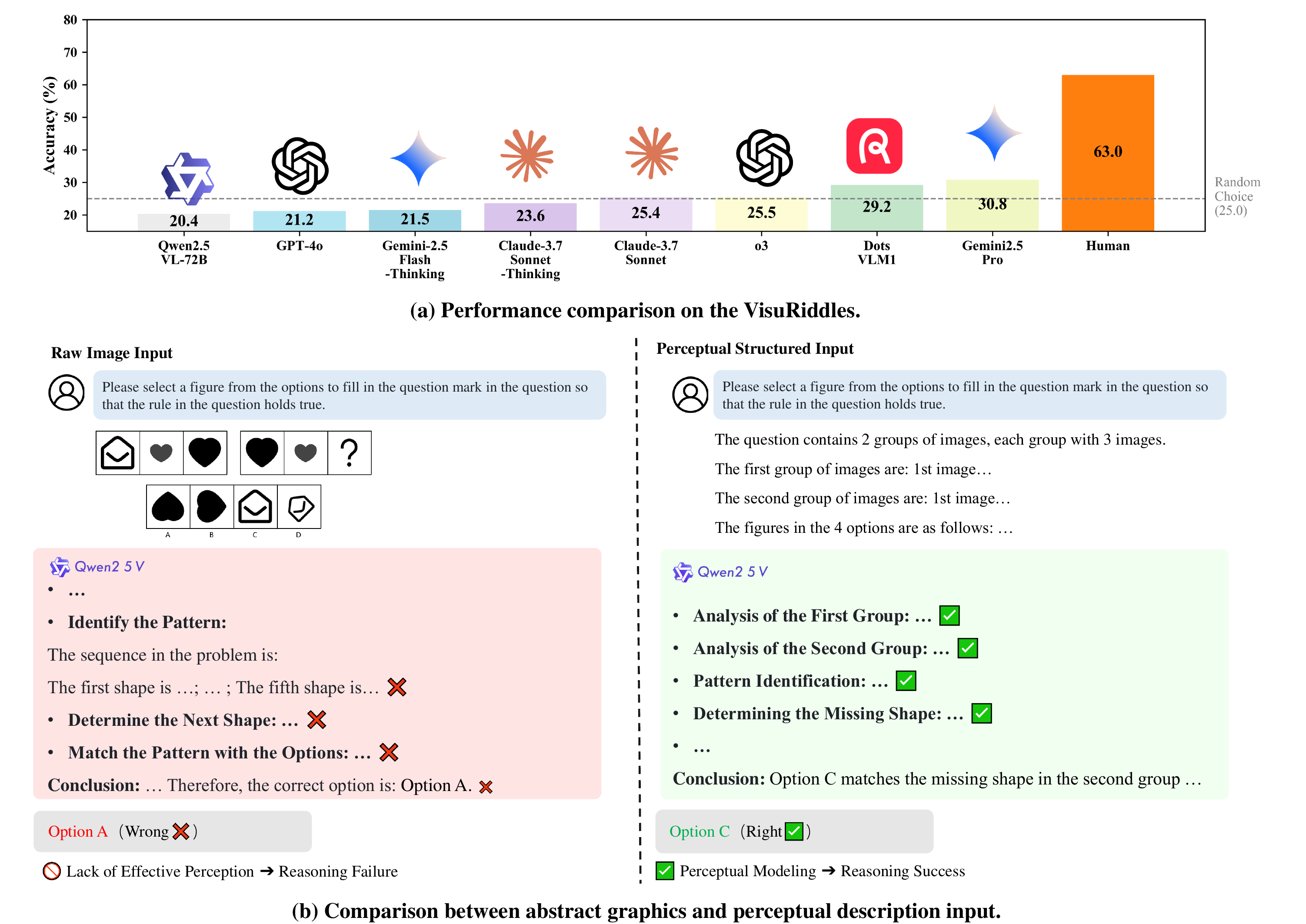}
  \caption{Performance and analysis of MLLMs on AVR. 
(a) Most advanced MLLMs achieve limited accuracy on VisuRiddles, often close to random choice and far below human performance. 
(b) Model responses to abstract graphics and their perceptual descriptions show that once equipped with perceptual capability, MLLMs can succeed in AVR task.}
  \label{fig:fig1}
\end{figure*}

\section{Introduction}
Reasoning is a core component of human intelligence~\citep{johnson2010mental}. Enhancing this capability in Multimodal Large Language Models (MLLMs) is key to bridging the gap to human-level performance~\citep{schulze2025visual,caffagni2024revolution}.
Recent advancements~\citep{bai2025qwen2,chen2024expanding,wu2024deepseek,yu2024texthawk,yu2024texthawk2,li2024monkey} in MLLMs have made significant progress in various general visual tasks, but still struggle with  Abstract Visual Reasoning (AVR) tasks~\citep{mitchell2311comparing}. 
As shown in Fig.~\ref{fig:fig1}(a), even advanced models such as Gemini2.5 Pro continue to face significant challenges in this task, showing a substantial gap compared to human performance. 
While such AVR tasks are relatively easy for humans, even the most advanced MLLMs struggle to solve them. It unexpected that models with strong performance in general understanding and reasoning tasks still fall short in handling such reasoning challenges. In fact, the challenges of AVR primarily stem from two aspects: fine-grained perception and reasoning. Compared to the recent advances in enhancing reasoning capabilities, the ability to perceive subtle visual structures, such as position, style, and attribute, has been largely overlooked in existing studies~\citep{xu2025visulogic, jiang2024marvel}. Unlike humans, which naturally process such fine-grained visual cues, MLLMs often lack the perceptual ability required for abstract graphics understanding~\citep{tong2024eyes, cao2024visual}. As illustrated in Fig.~\ref{fig:fig1}(b), once abstract graphics are reformulated into structured perceptual descriptions, the model is able to generate accurate responses. This highlights that strengthening perceptual capability is a crucial factor for advancing AVR. Therefore, genuine progress in enabling MLLMs to master AVR requires addressing both perception and reasoning, with special emphasis on enhancing the often-overlooked perceptual capability.

To advance the study of AVR tasks, we introduce the VisuRiddles benchmark, which is primarily derived from real riddles and enables objective evaluation of MLLMs’ performance. Since existing datasets lack fine-grained perceptual annotations, we further design the VisuRiddles Synthesizer, which automatically generates AVR instances enriched with structured perceptual descriptions. Leveraging this synthesized data, we perform Supervised Fine-Tuning (SFT) to enhance the model’s ability to perceive fine-grained visual cues, thereby equipping it with the perceptual foundation necessary for AVR. 
Although SFT substantially improves the perceptual capability of MLLMs, they still face persistent bottlenecks in AVR, including errors in perceptual strategy (e.g., difficulty in determining whether the appropriate interpretive cue lies in the figure’s symmetry axis or in its right-angle structures) and limited reasoning ability on more difficult instances. Therefore, we introduce Reinforcement Learning (RL), which leverages the enhanced perceptual capability to guide models toward more reliable perceptual grounding and further strengthen their reasoning ability.

Building upon these components, we develop the Perception-Augmented Visual Reasoner (PAVR), which unifies enhanced perceptual capability with improved reasoning ability to address the dual challenges of AVR. Experimental results show that PAVR significantly outperforms advanced commercial models on AVR tasks, highlighting a systematic solution to AVR tasks. 

Our contributions can be summarized as follows: 

(1) We introduce VisuRiddles, which comprises a benchmark derived from real riddles for objectively evaluating MLLMs on AVR tasks, and a synthesizer that automatically generates AVR instances enriched with structured perceptual descriptions.

(2) We leverage synthesized data for SFT to enhance fine-grained perceptual capability for AVR tasks, and further adopt RL to improve reasoning ability and stabilize perceptual strategy selection, resulting in the Perception-Augmented Visual Reasoner.

(3) We conduct extensive experiments to demonstrate our key findings, including the reasoning limitations of current MLLMs, the effectiveness of our synthesized riddles, and the effective approach to addressing abstract visual problems.

\section{Related Work}

\subsection{Multimodal Benchmarks}

High-quality benchmarks are essential for driving progress in MLLMs. Earlier evaluation effort predominantly centered on basic visual comprehension tasks, leading to the development of extraction-oriented benchmarks~\citep{mathew2021docvqa,masry2022chartqa,liu2024ocrbench}. These benchmarks primarily assess a model’s ability to identify and align explicit visual elements. However, they depend heavily on surface-level cues, offering little assessment of an MLLM’s capability for abstraction and logical reasoning. 
Therefore, general-purpose benchmarks~\citep{xu2023mmbench,bitton2023visit,yang2024mathglm, yue2024mmmupro,yue2024mmmu} are widely focused on. These benchmarks span multiple tasks, such as visual understanding and mathematical reasoning, and are designed to assess the generalist capabilities of MLLMs. However, they tend to emphasize knowledge-based evaluation over visual logical reasoning. Moreover, due to their strong semantic dominance, MLLMs may directly rely on their inherent knowledge, making it difficult to disentangle and evaluate models’ true visual reasoning capabilities.
Therefore, some studies have shifted toward evaluating visual logical reasoning, leading to logic-oriented benchmarks such as RAVEN~\citep{zhang2019raven}, I-RAVEN~\citep{hu2021stratified}, RAVEN-FAIR~\citep{benny2021scale}, CVR~\citep{zerroug2022benchmark}, RPMs~\citep{zhang2024far}, MaRs-VQA~\citep{cao2024visual}, MathVerse~\citep{zhang2024mathverse}, MARVEL~\citep{jiang2024marvel}, PuzzleVQA~\citep{chia2024puzzlevqa}, VisCogBench~\citep{cao2024visual}, VisuLogic~\citep{xu2025visulogic}, VisualPuzzle~\citep{song2025visualpuzzles}, VisualSphinx~\citep{feng2025visualsphinx} and MV-MATH~\citep{wang2025mv}. These benchmarks use structured diagrams, mathematical forms, and spatial layouts to assess analogical reasoning, compositionality, and pattern induction, signaling a move toward higher-level AVR. However, these benchmarks still exhibit notable limitations, particularly in their partial reliance on external knowledge and lack breadth in reasoning coverage.

\subsection{Multimodal Large Language Models for Visual Reasoning}

Advanced commercial LLMs, including the GPT~\citep{chatgpt}, Gemini~\citep{gemini}, and Claude~\citep{claude3} series, have exhibited impressive capabilities in multimodal understanding and generation. These advances have spurred systematic research on MLLMs in both academia and industry. Early representative general-purpose MLLMs, such as LLaVA~\citep{liu2023visual}, Instruct-BLIP~\citep{instructblip}, Qwen-VL~\citep{bai2023qwen}, and Intern-VL~\citep{chen2024internvl}, perform well on general multimodal tasks but exhibit limitations in fine-grained visual perception. Subsequently, DeepSeek-VL~\citep{lu2024deepseek}, Monkey~\citep{li2024monkey}, TextMonkey~\citep{textmonkey}, InternLM-XComposer2-4KHD~\citep{dong2024internlmxcomposer24khdpioneeringlargevisionlanguage}, Qwen2-VL~\citep{wang2024qwen2vlenhancingvisionlanguagemodels} have introduced high-resolution visual perception mechanisms, establishing a more robust foundation for fine-grained visual understanding and reasoning. However, such improvements primarily expand the perceptual scope of MLLMs without substantially enhancing their reasoning capabilities. Therefore, inference-time scaling methods including MMVP~\citep{tong2024eyes}, LLaVA-CoT~\citep{xu2024llavacot}, BBA~\citep{zhao2024bba}, R-CoT~\citep{deng2024r}, and RedStar~\citep{xu2025redstar} have introduced and advanced inference-time scaling techniques such as chain-of-thought (CoT), significantly enhancing the visual reasoning capabilities of MLLMs. More recently, following the advances of Kimi K1.5~\citep{team2025kimi} and DeepSeek-R1~\citep{guo2025deepseek}, a series of studies, such as R1-OneVision~\citep{yang2025r1}, LMM-R1~\citep{peng2025lmm}, MM-EUREKA~\citep{meng2025mm}, R1-V~\citep{chen2025r1v}, Visual-RFT~\citep{liu2025visual}, VisualPRM~\citep{wang2025visualprm}, OThink-MR1~\citep{liu2025othink}, and VLM-R1~\citep{shen2025vlm}, have employed reinforcement learning to develop adaptive reasoning policies and improve generalization in complex visual reasoning tasks.

\section{VisuRiddles}
In this section, we introduce \textbf{VisuRiddles}, a innovative and comprehensive resource for AVR research. It consists of the \textbf{VisuRiddles Benchmark} (Sec.~\ref{exp:Benchmark}), built from real-world problems to objectively evaluate MLLMs’ AVR performance, and the \textbf{VisuRiddles Synthesizer} (Sec.~\ref{exp:Synthesis}), designed to generate diverse perceptually annotated AVR instances for model training. 

\subsection{VisuRiddles Benchmark}
\label{exp:Benchmark}

\begin{figure*}[htbp]
  \centering
  \includegraphics[width=\linewidth]{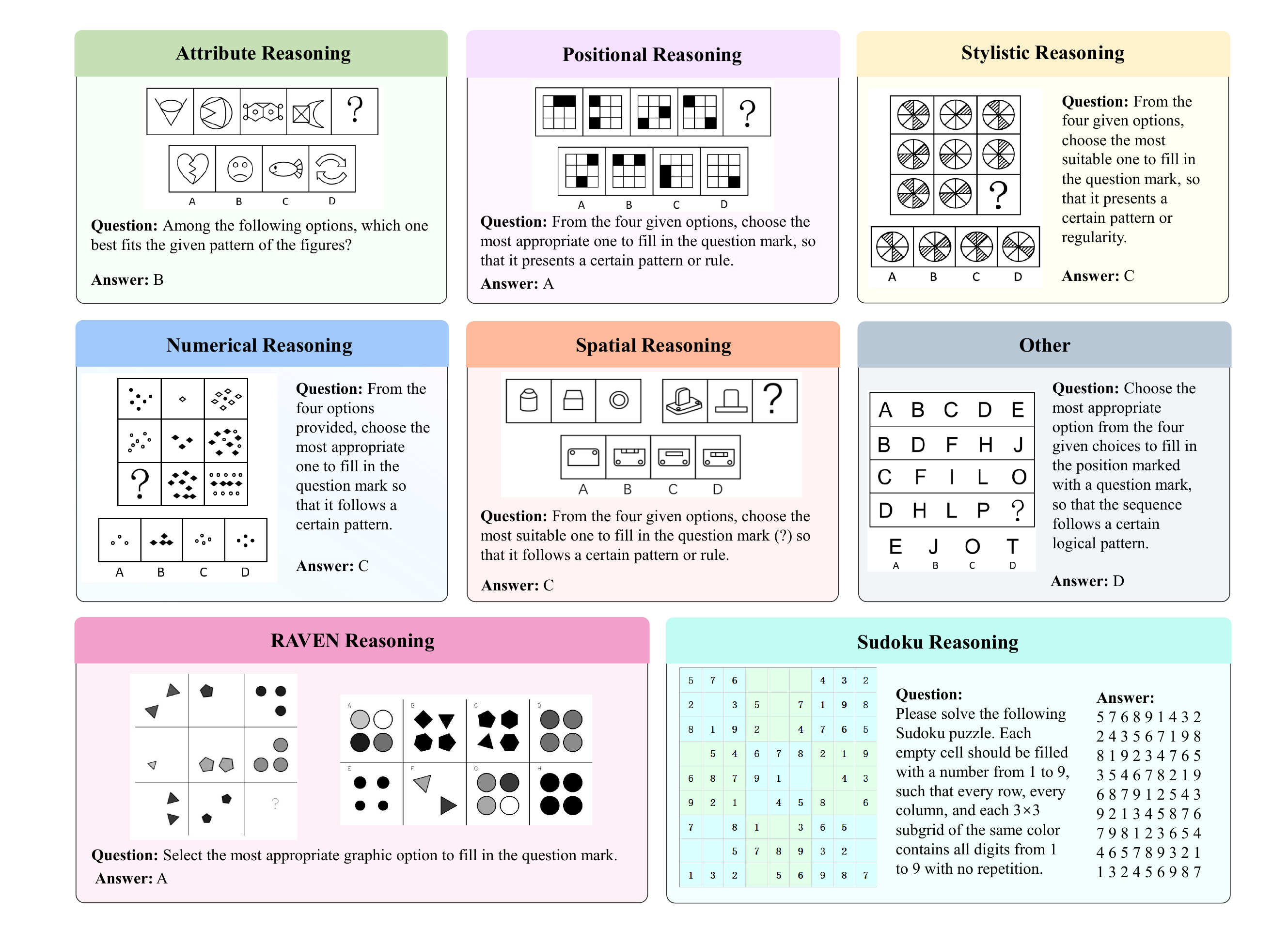}
  \caption{Representative examples from VisuRiddles. The benchmark includes eight reasoning categories, designed to comprehensively evaluate diverse reasoning capabilities of MLLMs.}
  \label{fig:appendix_figure1}
\end{figure*}

Unlike RAVEN~\citep{zhang2019raven} and PuzzleVQA~\citep{chia2024puzzlevqa}, which focus on general visual reasoning tasks, we target the most challenging part of AVR tasks. Thus, the VisuRiddles benchmark has been constructed to evaluate MLLMs. It explicitly evaluates models across five key dimensions: a) \textbf{Numerosity} assesses the model’s ability to perceive and reason about quantity and distribution; b) \textbf{Attribute} evaluates the understanding of intrinsic visual features that determine structural semantics; c) \textbf{Style} tests the capability to identify and generalize transformation-based visual patterns; d) \textbf{Position} reflects the ability to reason over the relative positions and layout of visual elements; e) \textbf{Spatiality} examines the understanding of three-dimensional structures, shape variations, and spatial transformations in abstract graphics. To extend the evaluation beyond low-level perceptual reasoning and address limitations in task diversity, structural complexity, and the inherent uncertainty introduced by single-choice formats, we additionally incorporate two high-level reasoning tasks that require analogical abstract reasoning and consistency-based logical reasoning, as represented by \textbf{RAVEN} and \textbf{Sudoku}, respectively. Finally, VisuRiddles also includes a subset, namely Other of AVR tasks involving planar shape composition and character-based semantic patterns. VisuRiddles provides a unified benchmark spanning basic to high-level reasoning, enabling comprehensive assessment of AVR, with representative samples shown in Fig.~\ref{fig:appendix_figure1}. We construct the VisuRiddles benchmark through a structured pipeline comprising \textbf{Collection}, \textbf{Cleaning}, and \textbf{Consolidation}. We provide the construction details in Appendix~\ref{appedix:benchmark}.

We initially collected 1,275 AVR problems, each accompanied by expert-level analysis and answer. After careful filtering and quality control, we construct VisuRiddles with 800 samples in the basic categories and 200 samples in the high-level categories. Key statistics of VisuRiddles are summarized in Tab.~\ref{statistic}. For the basic AVR tasks, each question is formulated as a single-choice problem, with answer option distributions of A (25.75\%), B (24.75\%), C (24.375\%), and D (25.125\%). In the high-level categories, RAVEN Reasoning and Sudoku Reasoning each account for 10\% of the total data, and require models to generate exact symbolic outputs to be considered correct.

\begin{table*}[t!]
\caption{The statistics of VisuRiddles, including the number questions, average question length (in tokens), number of associated images, and the corresponding answer format.}
  \label{statistic}
  \centering
  \begin{tabular}{lcccc}
    \toprule
    \cmidrule(r){1-2}
    Type     &    Questions Number  &Question length & Answer Formats   \\
    \toprule
    Sudoku     & 100  & 48.3  & Constraint-based grid\\
    RAVEN      & 100  & 70.7  & Single-choice  \\
    Numerical  & 250  & 26.3  & Single-choice \\
    Stylistic  & 117  & 26.3  & Single-choice \\
    Attribute  & 97   & 27.0  & Single-choice\\
    Positional & 111  & 25.5  & Single-choice \\
    Spatial    & 156  & 27.2  & Single-choice \\
    Other      & 69   & 30.3  & Single-choice \\
    Total               & 1000 & 32.5  & - \\    
    \bottomrule
  \end{tabular}
\end{table*}

\begin{figure*}[htbp]
  \centering
  \includegraphics[width=1\textwidth]{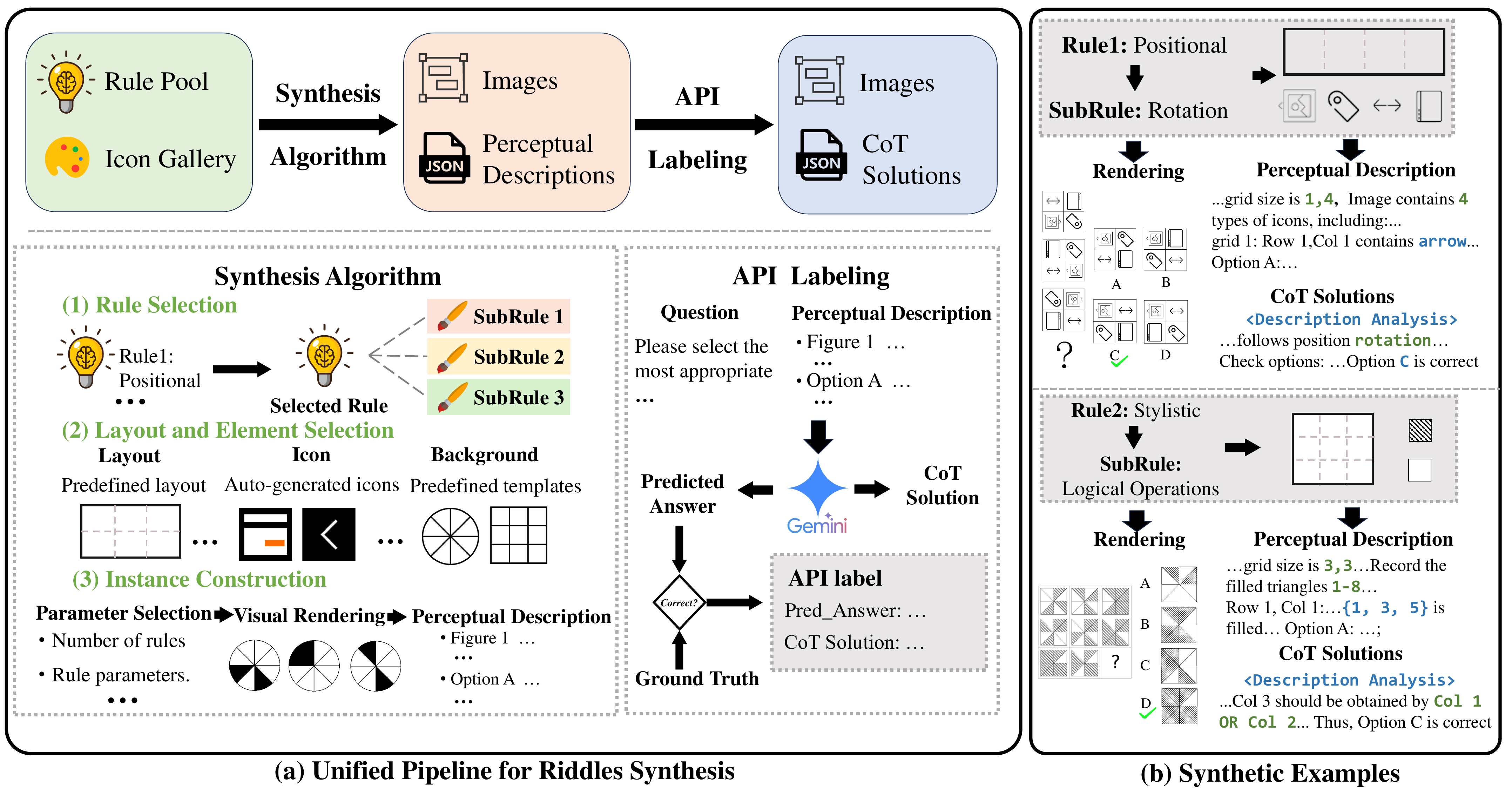}
  \caption{ Overview of the \textbf{VisuRiddles Synthesizer}. (a) A unified pipeline for generating abstract graphics with fine-grained perceptual descriptions. (b) Visualization of synthesized riddles based on positional rule and stylistic rule.}
  \label{synth pipeline}
\end{figure*}

\subsection{VisuRiddles Synthesizer}
\label{exp:Synthesis}
Recent MLLMs demonstrate strong general reasoning, but their limited performance on AVR is largely attributed to insufficient capability of fine-grained perception for abstract graphics. Most existing datasets lack intermediate perceptual description, providing only question-answer pairs~\citep{xu2025visulogic, jiang2024marvel, song2025visualpuzzles}. This prevents explicit modeling of the perception-to-reasoning process, leading to black-box inference, weak inductive capability, and poor generalization. More critically, existing commercial models struggle to annotate perceptual processes effectively, and manual labeling requires significant expert effort, both of which further exacerbate the issue. To address this, we introduce \textbf{VisuRiddles Synthesizer}, a Riddles Synthesis Framework to provides abstract graphics with aligned fine-grained perceptual description.

Figure~\ref{synth pipeline} illustrates the pipeline of our VisuRiddles Synthesizer for AVR task. The VisuRiddles Synthesizer pipeline comprises two stages: Riddles Construction and API Labeling. The former generates abstract visual instances with aligned perceptual descriptions, while the latter generates reasoning chains based on these descriptions. Details and representative examples are provided in Appendix~\ref{appendix:synthesis} and Appendix~\ref{appendix:synthesis example}.

The VisuRiddles Synthesizer generates data that spans five core perceptual reasoning types and two high-level reasoning tasks. The detailed configurations of VisuRiddles Synthesizer are provided in Table~\ref{tab:appendix_table1}. The VisuRiddles Synthesizer not only provides riddles' visual instances but also structured perceptual descriptions, thereby enhancing the model’s fine-grained perceptual ability for AVR tasks. It is noteworthy that the riddles instances generated by the VisuRiddles Synthesizer focus on improving fine-grained perception, rather than reasoning. Therefore, their reasoning difficulty is deliberately kept lower than that of real-world riddles.

\section{Perception-Augmented Visual Reasoner}

\begin{figure*}[t!]
  \centering
  \includegraphics[width=1\textwidth]{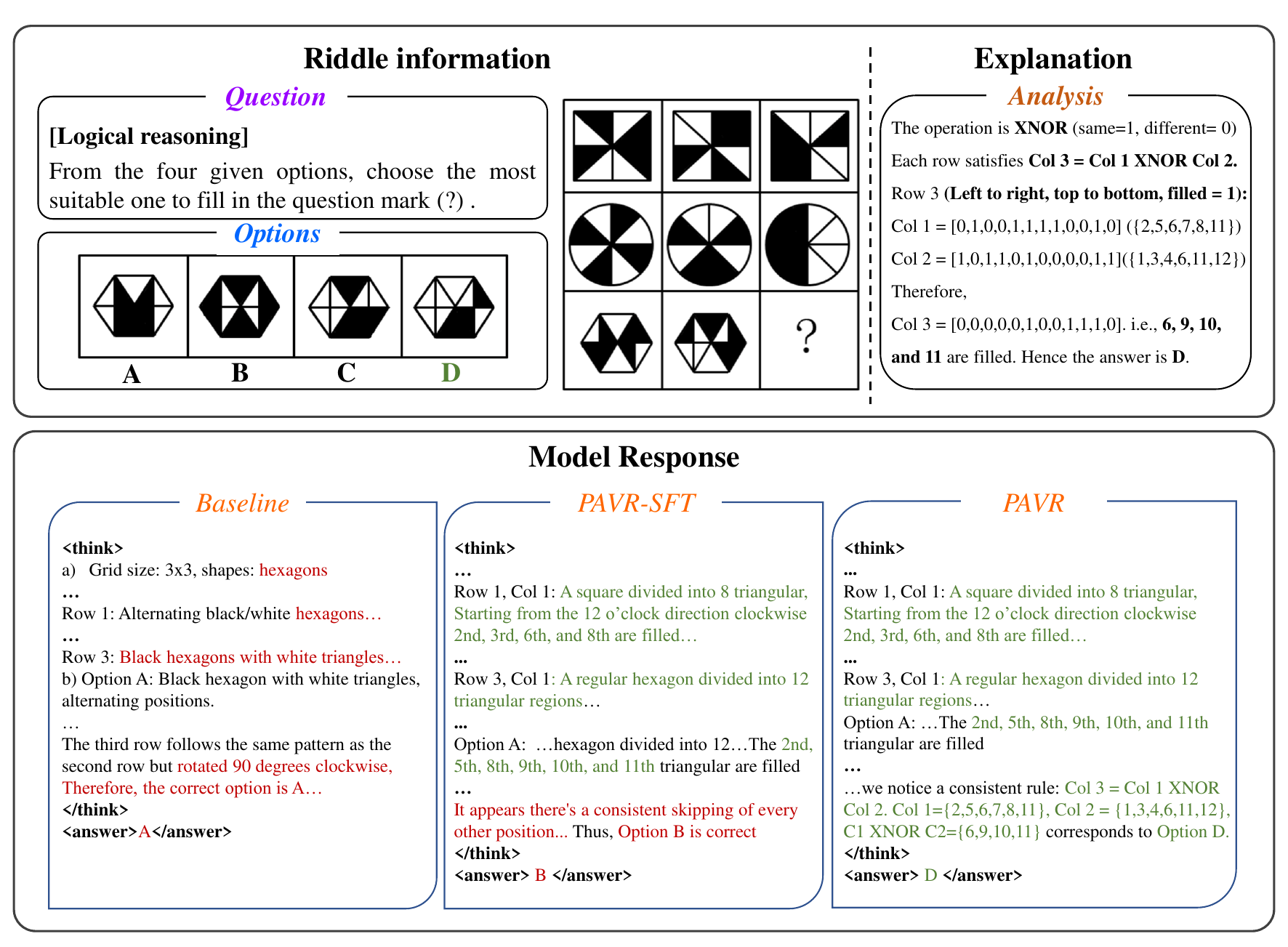}
  \caption{Overview of PAVR. (i) Baseline: Incorrect perception leads to incorrect results. (ii) PAVR-SFT, which is trained on synthesized data to enhance fine-grained perceptual ability, can accurately understand visual content in riddles but fails in pattern identification. (iii) PAVR, which builds upon PAVR-SFT with reinforcement learning, can effectively recognize patterns and derive the correct answer.}
  \label{fig:placeholder}
\end{figure*}

To tackle the challenges of AVR, we introduce the Perception-Augmented Visual Reasoner (PAVR). As illustrated in Fig.~\ref{fig:placeholder}, PAVR is developed through a two-stage training paradigm. In the first stage, SFT equips the model with the ability to capture fine-grained visual cues in abstract graphics. Building upon this, the second stage employs RL to effectively integrate these visual cues into stable reasoning strategies for solving complex tasks, resulting in PAVR. These two stages are complementary: SFT enables the model to see clearly, while RL guides it to reason reliably.

\textbf{Perceptual Augment via Supervised Fine-Tuning.}
The lack of fine-grained perceptual ability remains a major bottleneck for MLLMs on AVR tasks, yet existing real-world datasets typically provide only single-choice answer and rarely provide explicit perceptual annotations to supply sufficient supervision. To address this limitation, we employ the \textit{VisuRiddles Synthesizer} described in Sec.~\ref{exp:Synthesis} to generate AVR instances across seven categories, each enriched with structured perceptual descriptions. These synthesized examples serve as training data for SFT, enhancing the model’s perceptual ability and enabling it to capture fine-grained cues in abstract graphics. This establishes a reliable foundation for subsequent reasoning optimization through reinforcement learning.  

\textbf{Reasoning Optimization via Reinforcement Learning.}
Although SFT on synthesized data improves fine-grained perceptual ability, it still faces limitations in AVR, including unstable perceptual strategy selection (e.g., axis vs. angles) and insufficient reasoning on complex tasks. Therefore, we apply Group Relative Policy Optimization (GRPO)~\citep{guo2025deepseek} to endow the model with more reliable perceptual grounding and stronger reasoning ability. The reward design includes: (i) Answer Reward, giving 1 for correct answers and 0 otherwise, and (ii) Format Reward, incentivizing outputs that match the required template: <think> ... </think> <answer> ... </answer>. The training data for GRPO is likewise generated by the \textit{VisuRiddles Synthesizer}. This process enables the model to achieve more reliable perceptual grounding and stronger reasoning capabilities. 

\section{Experiment}

\subsection{Experimental Setup}

We use Qwen2.5-VL-7B~\citep{bai2025qwen2} as the baseline model to develop PAVR. In the supervised fine-tuning stage, the model is trained for 20 epochs using 20K synthesized AVR instances generated by the VisuRiddles Synthesizer. Subsequently, in the GRPO stage, optimization is performed using 4K synthesized instances over 40 epochs. All experiments are conducted on a setup comprising 8 NVIDIA A800 80G GPUs. Evaluation settings are summarized in Appendix~\ref{evaluated_models}. 

\begin{table*}[t!]
\caption{Evaluation result on VisuRiddles benchmark. The superscript number indicates the number of answer options (e.g., ``4'' means a 4-choice question with one correct option; * represents a vast combinatorial solution space). The best results are marked \textbf{bold} and the second results are \underline{underlined}.}
\label{tab:main_res}
\begin{tabularx}{\textwidth}{l|c|
>{\centering\arraybackslash}X  
>{\centering\arraybackslash}X
*{4}{>{\centering\arraybackslash}X}
@{\hspace{15pt}}
*{3}{>{\centering\arraybackslash}X}
}
\toprule
    Model & Param & Num\textsuperscript{4} & Styl\textsuperscript{4}  & Attr\textsuperscript{4} & Posit\textsuperscript{4} & Spat\textsuperscript{4} &Sudo\textsuperscript{*} & Rav\textsuperscript{8}& Other\textsuperscript{4}& Avg \\
    \midrule
    Human & - & 61.3 & 60.9 & 67.5 & 67.9 & 58.8 & - & - & 61.9 & -\\
    \hline
    \multicolumn{11}{c}{\textit{Open-Source MLLMs}}\\ \midrule 
    Minicpm-V-2.6 & 8B & 21.6 & 26.5 & 23.7 & 31.5 & 23.1 & 0.0 & 10.0 & 24.6 & 20.6 \\
    InternVL2.5-8B & 8B &21.2 & 13.7 & 30.9 & 23.4 & 23.1 & 0.0 & 0.0 & 29.0 & 18.1 \\
    InternVL2.5-8B-MPO & 8B & 22.0 & 23.9 & 27.8 & 25.2 & 21.8 & 0.0 & 3.0 & 27.5 & 19.4 \\
    Deepseekvl2 & 27B & 21.2 & 27.4 & 18.6 & 18.0 & 17.3 & 0.0 & 15.0 & 13.0 & 17.4 \\
    InternVL2.5-38B & 38B & 22.0 & 19.7 & 29.9 & 26.1 & 29.5 & 1.0 & 19.0 & 30.4 & 22.3 \\
    InternVL2.5-38B-MPO & 38B & 26.0 & 20.5 & 19.6 & 27.0 & 27.6 & 0.0 & 18.0 & 31.9 & 22.1 \\
    Qwen2.5VL-32B & 32B & 23.2 & 20.5 & 30.9 & 27.0 & 29.5 & 0.0 & 40.0 & 24.6 & 24.5 \\
    InternVL2.5-78B & 78B & 26.0 & 25.6 & 27.8 & 29.7 & 23.7 & 0.0 & 16.0 & 27.5 & 22.7 \\
    InternVL2.5-78B-MPO & 78B & 25.6 & 30.8 & 24.7 & 27.9 & 26.3 & 0.0 & 12.0 & 20.3 & 22.2 \\
    InternVL2.5-78B-MPO(cot) & 78B & 23.2 & 23.1 & 24.7 & 28.8 & 22.4 & 0.0 & 11.0 & 23.2 & 20.3 \\
    Qwen2.5VL-72B & 72B & 23.6 & 23.1 & 19.6 & 30.2 & 26.9 & 0.0 & \underline{62.0} & 23.9 & 25.9 \\
    Qwen2.5VL-72B(cot) & 72B & 27.2 & 26.5 & 24.7 & 19.8 & 28.2 & 0.0 & 55.0 & 23.2 & 26.0 \\
    Qwen3-VL-235B-A22B-Instruct & 235B & 31.2 & 28.2 & 36.1 & 34.2 & 26.3 & 29.0 & 53.0 & 27.5 & 32.6 \\
    Qwen3-VL-235B-A22B-Thinking & 235B & 31.2 & 29.9 & \underline{44.3} & 33.3 & 30.1 & 33.0 & 49.0 & 39.1 & 34.9\\
    Dots.vlm1 & 671B & 30.4&31.6&38.1&\underline{38.7}&35.3&1.0&20.0&33.3&29.2 \\
    \hline
    \multicolumn{11}{c}{\textit{Commercial products}}\\ \midrule 
    GPT-4o & - & 23.2& 21.4& 29.9& 27.9& 28.2& 0.0& 14.0& 24.6& 21.8 \\
    GPT-4o(cot) & -& 22.4& 23.9& 32.0& 29.7& 30.1& 0.0& 22.0& 29.0& 23.7 \\
    o3 & - & 28.4& \textbf{40.2}& 41.2& 27.0& 21.8& 0.0& 25.0& 33.3& 27.0 \\
    claude-3-7-sonnet & -& 24.0 & 27.4 & 27.8 & 29.7 & 28.2 & 4.0 & 43.0 & 31.9 & 26.5 \\
    claude-3-7-sonne(cot) & -&24.4 & 36.8 & 27.8 & 29.7 & 25.6 & 3.0 & 36.0 & 27.5 & 26.2\\
    claude-3-7-sonnet-thinking &-& 24.0 & 24.8 & 33.0 & 23.4 & 25.6 & 3.0 & 24.0 & 26.1 & 23.2 \\
    Gemini-2.5-flash-thinking &-& 21.2 & 30.8 & 27.8 & 21.6 & 16.7 & 17.0 & 16.0 & 23.2 & 21.5\\
    Gemini2.5-pro & - & \underline{31.6} & 31.6 & 48.5 & 26.1 & 30.1 & 39.0 & 30.0 & \underline{44.9} & 33.9 \\
    GPT-5 & - & 30.8 & 30.8 & 38.1 & 32.4 & 30.8 & 2.0 & 29.0 & 31.9 & 28.7 \\
    \hline
    \multicolumn{11}{c}{\textit{Ours}}\\ \midrule 
    Baseline(Qwen2.5VL-7B) & 7B & 24.4 & 28.2 & 23.7 & 22.5 & 25.0 & 0.0 & 48.0 & 24.6 & 24.6 \\
    PAVR-SFT & 7B & 31.2 & 31.6 & \underline{44.3} & 31.5& \underline{45.5} & \underline{43.0} & 61.0 & 39.1 & \underline{39.5} \\ 
    PAVR & 7B & \textbf{39.6} & \underline{39.3} & \textbf{50.5} & \textbf{39.6} & \textbf{51.9} & \textbf{46.0} & \textbf{65.0} & \textbf{55.1} & \textbf{46.8} \\ 
    \bottomrule
\end{tabularx}
\end{table*}

\subsection{Main Results}
\label{exp:main_res}

Table~\ref{tab:main_res} presents performance of various MLLMs on VisuRiddles benchmark. We summarize the key observations as follows:

\textbf{Performance of Open-Source MLLMs.}
Most open-source MLLMs demonstrate limited performance on VisuRiddles, with accuracies across core categories often close to random choice and consistently failing on the two high-level reasoning tasks, which reveals a substantial gap between current model capabilities and human performance in AVR. Notably, the newly released Qwen3-VL-235B-A22B-Instruct and Qwen3-VL-235B-A22B-Thinking exhibit markedly stronger results, though such improvements remain exceptions among open-source models.

\textbf{Performance of Commercial Models with Thinking Mode.}
Several commercial MLLMs with ‘thinking’ mode achieve comparatively improved performance on VisuRiddles. Their advantage is particularly notable in structured tasks like Sudoku and Raven. These results indicate that AVR can benefit from advanced reasoning capabilities. However, the overall performance remains limited.

\textbf{Effectiveness of Model Scaling and CoT Prompting.}
Strategies such as increasing model size and applying CoT prompting, though commonly used to enhance reasoning, are ineffective for AVR tasks. Models with larger parameter scales do not consistently outperform smaller ones, and CoT yields only limited improvements on select tasks. These findings indicate that scaling in model parameter or inference time is insufficient to address the core challenges of AVR.

\textbf{Effectiveness of PAVR.}
PAVR significantly outperforms other MLLMs across all categories of VisuRiddles, especially in high-level reasoning tasks, demonstrating the model's ability to handle complex visual patterns. Notably, compared to the baseline, the significant improvement achieved by training on data with perceptual descriptions through SFT highlights the critical role of fine-grained perceptual ability in AVR tasks. On the other hand, the application of RL further enhances performance across various dimensions on VisuRiddles, with the improvements primarily stemming from enhanced reasoning capabilities through strategic guidance. 

We also provide the results of PAVR on the VisuLogic~\citep{xu2025visulogic} benchmark in ~\ref{appendix:visulogic}. The experimental results further support our conclusions.

\subsection{Visualization Results}
\label{exp:vis}
\begin{figure*}[t!]
  \centering
  \includegraphics[width=1\textwidth]{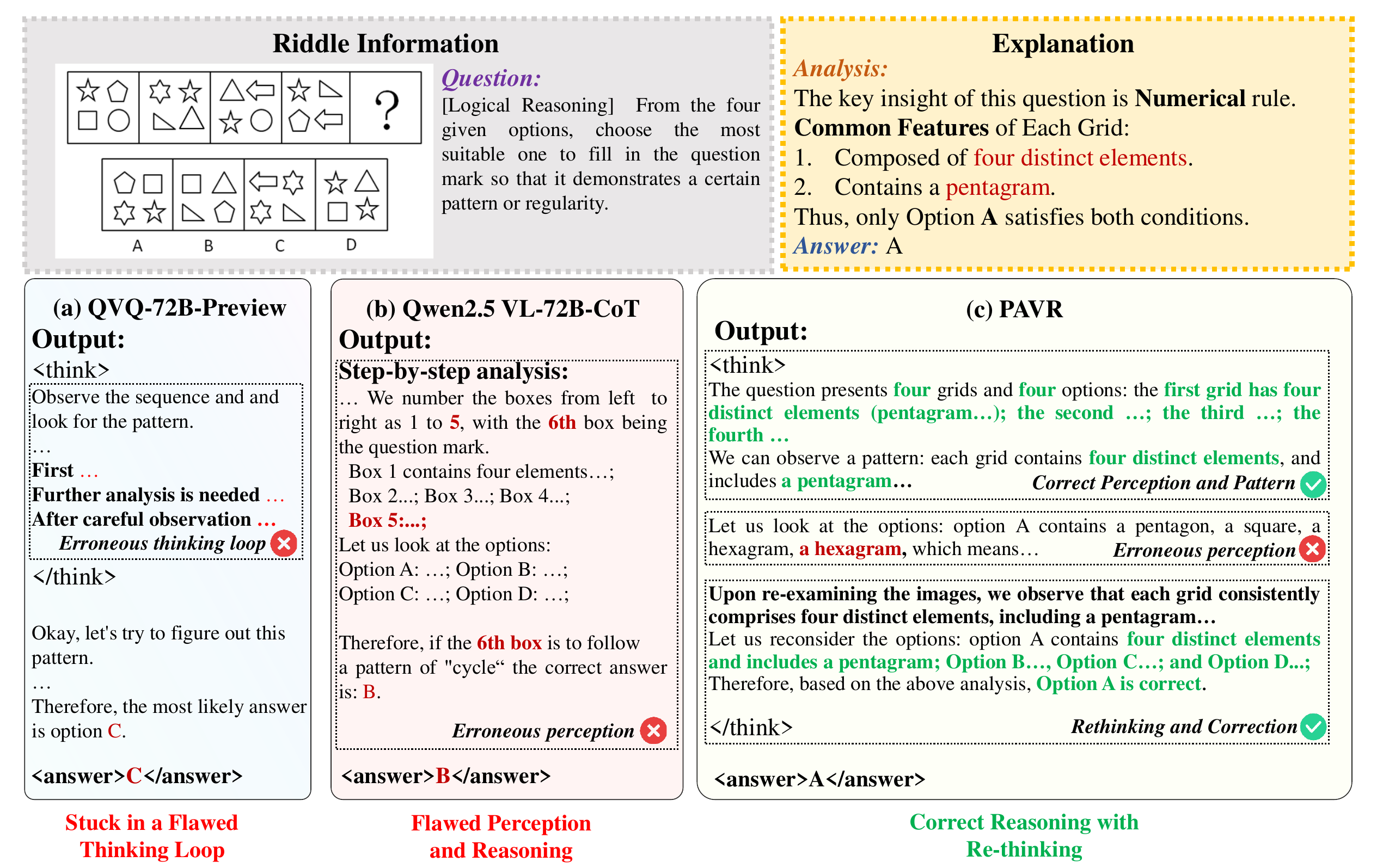}
  \caption{Case study comparing different reasoning strategies on a VisuRiddles example. (a) QVQ-72B reflects a flawed loop under the "thinking" mode. (b) Qwen2.5-VL-72B with CoT prompting exhibits incorrect perceptual understanding. (c) PAVR exhibits accurate perception and coherent reasoning, ultimately arriving at the correct answer.}
  \label{example}
\end{figure*}

To further analyze the reason that MLLMs underperform on AVR task, we examine two widely adopted reasoning enhancement strategies: inference-time scaling and CoT prompting. We conduct experiments on QVQ-72B-Preview (equipped with a "think" mode), Qwen2.5-VL-72B (with CoT prompting), and PAVR. 
Fig.~\ref{example} visualizes a representative case of a broadly observed phenomenon uncovered during our experiments. The results reveal clear differences in the reasoning behaviors across these models. The inference-time scaling approach, which leverages "think"-style prompting without any perceptual grounding, frequently leads to verbose yet logically inconsistent reasoning, often ending in incorrect answers. The CoT prompting strategy often suffers from perceptual errors, which cascade into flawed reasoning and ultimately lead to incorrect answers. In contrast, by grounding reasoning in fine-grained perceptual capability, PAVR correctly interprets the riddle and identifies the underlying pattern, ultimately producing the correct answer. Notably, while PAVR occasionally exhibits erroneous perception, we observe a ‘rethink’ phenomenon, where the model re-examines and corrects them, realigning the reasoning trajectory to reach the correct solution.

These results underscore the importance of perception in AVR. Inference-time scaling and CoT prompting alone cannot compensate for the lack of perceptual grounding, which often leads to incorrect or inconsistent reasoning. Conversely, the incorporation of fine-grained perceptual capability enables the model to capture subtle visual cues and sustain reliable performance on AVR tasks.

\subsection{Ablation Study}
\label{exp:abl}

\textbf{Ablation Study on Bottlenecks in AVR}. To investigate whether the performance gap in AVR stems from perceptual limitations rather than reasoning ability, we conduct an ablation study using our perception-annotated synthetic data. 
We evaluate two representative MLLMs, GPT-4o (closed-source) and Qwen2.5VL-72B (open-source), on a subset of our synthetic dataset under two input settings: (i) raw abstract graphics (V), and (ii) structured perceptual descriptions (P). 

The experimental results in Tab.~\ref{Perceptual Description} show that both models perform poorly when given only raw visual inputs. Taking Sudoku reasoning as an example, this task requires recognizing digit values and their positions within a grid, and inferring a globally consistent solution. Under the Visual setting, both GPT-4o and Qwen2.5VL-72B fail to handle such constraint-based reasoning, while in the perceptual descriptions setting, accuracy increases from near-zero to 15\% and 65\%, respectively. This improvement primarily stems from structured perceptual descriptions compensating for the MLLMs’ limited perceptual abstraction capabilities. The difficulty stems from the nature of Sudoku images: small digits, dense layout, and absent semantic cues make them hard for MLLMs to interpret, though humans process such structure effortlessly through rapid visual scanning.

\begin{table*}[htbp]
\caption{Impact of Descriptions vs. Visual Inputs in MLLMs. Where "V" denotes inputting abstract graphics directly, while “P” refers to replacing graphics with structured perceptual descriptions.} 
\centering
\resizebox{\textwidth}{!}{
\begin{tabular}{lllllllll}
\toprule
\textbf{Model}  &  Num. & Styl. & Attr. & Posit. & Spat. &Sudo. & Rav. & Avg   \\

\midrule

GPT-4o(V)         & 35.0 & 32.0 & 38.0 & 36.0 & 32.0 & 0.0   & 20.0 & 27.6 \\
GPT-4o(P)        & 62.0\textcolor{red}{\scriptsize{ (+27.0)}} & 53.0\textcolor{red}{\scriptsize{ (+21.0)}} & 80.0\textcolor{red}{\scriptsize{ (+42.0)}} & 68.0\textcolor{red}{\scriptsize{ (+32.0)}} & 100.0\textcolor{red}{\scriptsize{ (+68.0)}} & 15.0\textcolor{red}{\scriptsize{ (+15.0)}}  & 25.0\textcolor{red}{\scriptsize{ (+5.0)}} & 60.1\textcolor{red}{\scriptsize{(+32.5)}} \\
Qwen2.5VL(V)  & 41.0 & 43.0 & 50.0 & 32.0 & 40.0 & 0.0      & 10.0 & 30.9 \\
Qwen2.5VL(P) & 73.0\textcolor{red}{\scriptsize{(+32.0)}} & 83.0\textcolor{red}{\scriptsize{(+40.0)}} & 80.0\textcolor{red}{\scriptsize{(+30.0)}} & 79.0\textcolor{red}{\scriptsize{(+47.0)}} & 100.0\textcolor{red}{\scriptsize{(60.0)}} & 65.0\textcolor{red}{\scriptsize{(+65.0)}}  & 35.0\textcolor{red}{\scriptsize{(+25.0)}} & 73.6\textcolor{red}{\scriptsize{(+42.7)}} \\

\bottomrule
\end{tabular}
}
\label{Perceptual Description}

\end{table*}

\textbf{Ablation Study on Reasoning Enhancements in AVR}. To examine the contribution of reasoning-oriented enhancements, we conduct an ablation study on three components: Caption, CoT, and GRPO, as shown in Table~\ref{Reasoning Enhancement}. We can draw the following conclusions form the experiments: (i) perception serves as the essential foundation, as reasoning enhancements alone contribute only marginal improvements; (ii) relying solely on structured perceptual descriptions for SFT yields limited gains and weak generalization, whereas augmenting SFT with CoT annotations generated by MLLMs alleviates this issue, endowing the model with fine-grained perceptual ability and partial reasoning capability; and (iii) Building on a perception-grounded model, the incorporation of GRPO further unifies perception and reasoning, ultimately delivering substantial performance improvements.

\begin{table}[htbp]
\caption{Ablation study on Reasoning Enhancements. Where Caption denotes training with fine-grained perceptual descriptions, CoT denotes augmenting training with chain-of-thought reasoning traces that incorporate structured perceptual descriptions generated by MLLMs, and GRPO refers to applying Group Relative Policy Optimization to strength model’s reasoning capability.} 
\center
{\small{
\label{Reasoning Enhancement}
\begin{tabular}{lccl}
\toprule
\textbf{Model}  & SFT & RL & Avg   \\


\midrule
Baseline (Qwen2.5-VL)    & ✗ & ✗ &24.6\\
Baseline + Caption   & ✓ & ✗ & 33.3\textcolor{red}{\scriptsize{ (+8.7)}}\\
Baseline + GRPO   & ✗ & ✓ & 29.4\textcolor{red}{\scriptsize{(+4.8)}}\\
Baseline + CoT (PAVR-SFT) & ✓ & ✗ &39.5\textcolor{red}{\scriptsize{(+14.9)}}\\
Baseline + CoT + GRPO (PAVR) & ✓ & ✓ & 46.8\textcolor{red}{\scriptsize{ (+22.2)}}\\
\bottomrule
\end{tabular}
}
}
\end{table}

\section{Conclusion}

In this work, we observe that the primary bottleneck of MLLMs in AVR lies in their overlooked fine-grained perceptual capability. To address this issue, we introduce VisuRiddles, a unified resource that comprises a benchmark for objectively evaluating MLLMs on AVR tasks and a synthesizer for generating training instances with structured perceptual descriptions. Building on this resource, we propose PAVR, which unifies perception and reasoning through supervised fine-tuning for perceptual enhancement and GRPO-based optimization for reasoning improvement for AVR tasks. Experimental results demonstrate that current MLLMs perform poorly on AVR tasks. Moreover, scaling model parameters, CoT prompting, and inference-time scaling fail to effectively address this challenge. These findings highlight that fine-grained perception constitutes the fundamental bottleneck in AVR, while reasoning enhancements applied based on a perception-grounded model can further improve performance.

In short, our work not only offers a systematic resource for both evaluation and training, but also provides a practical framework for addressing the AVR challenge.

\textbf{Limitation:} Due to limitations in time and human labor, the scale of the resources (e.g., rule design, icon library) used for data synthesis is restricted, which somewhat limits the diversity and richness of the synthetized data. In future work, we plan to expand the resource pool to a larger scale with greater complexity, further enriching the synthesized riddles.

{
    \small
    \bibliographystyle{unsrt}
    \bibliography{main}

\begin{thebibliography}{10}

\bibitem{johnson2010mental}
Philip~N Johnson-Laird.
\newblock Mental models and human reasoning.
\newblock {\em Proceedings of the National Academy of Sciences}, 107(43):18243--18250, 2010.

\bibitem{cao2024visual}
Xu~Cao, Bolin Lai, Wenqian Ye, Yunsheng Ma, Joerg Heintz, Jintai Chen, Jianguo Cao, and James~M Rehg.
\newblock What is the visual cognition gap between humans and multimodal llms?
\newblock {\em arXiv preprint arXiv:2406.10424}, 2024.

\bibitem{yang2025r1}
Yi~Yang, Xiaoxuan He, Hongkun Pan, Xiyan Jiang, Yan Deng, Xingtao Yang, Haoyu Lu, Dacheng Yin, Fengyun Rao, Minfeng Zhu, et~al.
\newblock R1-onevision: Advancing generalized multimodal reasoning through cross-modal formalization.
\newblock {\em arXiv preprint arXiv:2503.10615}, 2025.

\bibitem{bai2025qwen2}
Shuai Bai, Keqin Chen, Xuejing Liu, Jialin Wang, Wenbin Ge, Sibo Song, Kai Dang, Peng Wang, Shijie Wang, Jun Tang, et~al.
\newblock Qwen2. 5-vl technical report.
\newblock {\em arXiv preprint arXiv:2502.13923}, 2025.

\bibitem{chen2024expanding}
Zhe Chen, Weiyun Wang, Yue Cao, Yangzhou Liu, Zhangwei Gao, Erfei Cui, Jinguo Zhu, Shenglong Ye, Hao Tian, Zhaoyang Liu, et~al.
\newblock Expanding performance boundaries of open-source multimodal models with model, data, and test-time scaling.
\newblock {\em arXiv preprint arXiv:2412.05271}, 2024.

\bibitem{wu2024deepseek}
Zhiyu Wu, Xiaokang Chen, Zizheng Pan, Xingchao Liu, Wen Liu, Damai Dai, Huazuo Gao, Yiyang Ma, Chengyue Wu, Bingxuan Wang, et~al.
\newblock Deepseek-vl2: Mixture-of-experts vision-language models for advanced multimodal understanding.
\newblock {\em arXiv preprint arXiv:2412.10302}, 2024.

\bibitem{yao2024minicpm}
Yuan Yao, Tianyu Yu, Ao~Zhang, Chongyi Wang, Junbo Cui, Hongji Zhu, Tianchi Cai, Haoyu Li, Weilin Zhao, Zhihui He, et~al.
\newblock Minicpm-v: A gpt-4v level mllm on your phone.
\newblock {\em arXiv preprint arXiv:2408.01800}, 2024.

\bibitem{li2024monkey}
Zhang Li, Biao Yang, Qiang Liu, Zhiyin Ma, Shuo Zhang, Jingxu Yang, Yabo Sun, Yuliang Liu, and Xiang Bai.
\newblock Monkey: Image resolution and text label are important things for large multi-modal models.
\newblock In {\em proceedings of the IEEE/CVF conference on computer vision and pattern recognition}, pages 26763--26773, 2024.

\bibitem{yu2024texthawk2}
Ya-Qi Yu, Minghui Liao, Jiwen Zhang, and Jihao Wu.
\newblock Texthawk2: A large vision-language model excels in bilingual ocr and grounding with 16x fewer tokens.
\newblock {\em arXiv preprint arXiv:2410.05261}, 2024.

\bibitem{schulze2025visual}
Luca~M Schulze~Buschoff, Elif Akata, Matthias Bethge, and Eric Schulz.
\newblock Visual cognition in multimodal large language models.
\newblock {\em Nature Machine Intelligence}, pages 1--11, 2025.

\bibitem{caffagni2024revolution}
Davide Caffagni, Federico Cocchi, Luca Barsellotti, Nicholas Moratelli, Sara Sarto, Lorenzo Baraldi, Marcella Cornia, and Rita Cucchiara.
\newblock The revolution of multimodal large language models: a survey.
\newblock {\em arXiv preprint arXiv:2402.12451}, 2024.

\bibitem{mathew2021docvqa}
Minesh Mathew, Dimosthenis Karatzas, and CV~Jawahar.
\newblock Docvqa: A dataset for vqa on document images.
\newblock In {\em Proceedings of the IEEE/CVF winter conference on applications of computer vision}, pages 2200--2209, 2021.

\bibitem{masry2022chartqa}
Ahmed Masry, Do~Xuan Long, Jia~Qing Tan, Shafiq Joty, and Enamul Hoque.
\newblock Chartqa: A benchmark for question answering about charts with visual and logical reasoning.
\newblock {\em arXiv preprint arXiv:2203.10244}, 2022.

\bibitem{liu2024ocrbench}
Yuliang Liu, Zhang Li, Mingxin Huang, Biao Yang, Wenwen Yu, Chunyuan Li, Xu-Cheng Yin, Cheng-Lin Liu, Lianwen Jin, and Xiang Bai.
\newblock Ocrbench: on the hidden mystery of ocr in large multimodal models.
\newblock {\em Science China Information Sciences}, 67(12):220102, 2024.

\bibitem{singh2019towards}
Amanpreet Singh, Vivek Natarajan, Meet Shah, Yu~Jiang, Xinlei Chen, Dhruv Batra, Devi Parikh, and Marcus Rohrbach.
\newblock Towards vqa models that can read.
\newblock In {\em Proceedings of the IEEE/CVF conference on computer vision and pattern recognition}, pages 8317--8326, 2019.

\bibitem{hsiao2022screenqa}
Yu-Chung Hsiao, Fedir Zubach, Gilles Baechler, Victor Carbune, Jason Lin, Maria Wang, Srinivas Sunkara, Yun Zhu, and Jindong Chen.
\newblock Screenqa: Large-scale question-answer pairs over mobile app screenshots.
\newblock {\em arXiv preprint arXiv:2209.08199}, 2022.

\bibitem{xu2023mmbench}
Cheng Xu, Xiaofeng Hou, Jiacheng Liu, Chao Li, Tianhao Huang, Xiaozhi Zhu, Mo~Niu, Lingyu Sun, Peng Tang, Tongqiao Xu, et~al.
\newblock Mmbench: Benchmarking end-to-end multi-modal dnns and understanding their hardware-software implications.
\newblock In {\em 2023 IEEE International Symposium on Workload Characterization (IISWC)}, pages 154--166. IEEE, 2023.

\bibitem{yue2024mmmu}
Xiang Yue, Yuansheng Ni, Kai Zhang, Tianyu Zheng, Ruoqi Liu, Ge~Zhang, Samuel Stevens, Dongfu Jiang, Weiming Ren, Yuxuan Sun, et~al.
\newblock Mmmu: A massive multi-discipline multimodal understanding and reasoning benchmark for expert agi.
\newblock In {\em Proceedings of the IEEE/CVF Conference on Computer Vision and Pattern Recognition}, pages 9556--9567, 2024.

\bibitem{bitton2023visit}
Yonatan Bitton, Hritik Bansal, Jack Hessel, Rulin Shao, Wanrong Zhu, Anas Awadalla, Josh Gardner, Rohan Taori, and Ludwig Schmidt.
\newblock Visit-bench: A benchmark for vision-language instruction following inspired by real-world use.
\newblock {\em arXiv preprint arXiv:2308.06595}, 2023.

\bibitem{yue2024mmmupro}
Xiang Yue, Tianyu Zheng, Yuansheng Ni, Yubo Wang, Kai Zhang, Shengbang Tong, Yuxuan Sun, Botao Yu, Ge~Zhang, Huan Sun, et~al.
\newblock Mmmu-pro: A more robust multi-discipline multimodal understanding benchmark.
\newblock {\em arXiv preprint arXiv:2409.02813}, 2024.

\bibitem{zhang2019raven}
Chi Zhang, Feng Gao, Baoxiong Jia, Yixin Zhu, and Song-Chun Zhu.
\newblock Raven: A dataset for relational and analogical visual reasoning.
\newblock In {\em Proceedings of the IEEE/CVF conference on computer vision and pattern recognition}, pages 5317--5327, 2019.

\bibitem{hu2021stratified}
Sheng Hu, Yuqing Ma, Xianglong Liu, Yanlu Wei, and Shihao Bai.
\newblock Stratified rule-aware network for abstract visual reasoning.
\newblock In {\em Proceedings of the AAAI Conference on Artificial Intelligence}, volume~35, pages 1567--1574, 2021.

\bibitem{benny2021scale}
Yaniv Benny, Niv Pekar, and Lior Wolf.
\newblock Scale-localized abstract reasoning.
\newblock In {\em Proceedings of the IEEE/CVF Conference on Computer Vision and Pattern Recognition}, pages 12557--12565, 2021.

\bibitem{barrett2018measuring}
David Barrett, Felix Hill, Adam Santoro, Ari Morcos, and Timothy Lillicrap.
\newblock Measuring abstract reasoning in neural networks.
\newblock In {\em International conference on machine learning}, pages 511--520. PMLR, 2018.

\bibitem{zerroug2022benchmark}
Aimen Zerroug, Mohit Vaishnav, Julien Colin, Sebastian Musslick, and Thomas Serre.
\newblock A benchmark for compositional visual reasoning.
\newblock {\em Advances in neural information processing systems}, 35:29776--29788, 2022.

\bibitem{huang2023language}
Shaohan Huang, Li~Dong, Wenhui Wang, Yaru Hao, Saksham Singhal, Shuming Ma, Tengchao Lv, Lei Cui, Owais~Khan Mohammed, Barun Patra, et~al.
\newblock Language is not all you need: Aligning perception with language models.
\newblock {\em Advances in Neural Information Processing Systems}, 36:72096--72109, 2023.

\bibitem{moskvichev2023conceptarc}
Arseny Moskvichev, Victor~Vikram Odouard, and Melanie Mitchell.
\newblock The conceptarc benchmark: Evaluating understanding and generalization in the arc domain.
\newblock {\em arXiv preprint arXiv:2305.07141}, 2023.

\bibitem{zhang2024far}
Yizhe Zhang, He~Bai, Ruixiang Zhang, Jiatao Gu, Shuangfei Zhai, Josh Susskind, and Navdeep Jaitly.
\newblock How far are we from intelligent visual deductive reasoning?
\newblock In {\em COLM}, 2024.

\bibitem{lu2023mathvista}
Pan Lu, Hritik Bansal, Tony Xia, Jiacheng Liu, Chunyuan Li, Hannaneh Hajishirzi, Hao Cheng, Kai-Wei Chang, Michel Galley, and Jianfeng Gao.
\newblock Mathvista: Evaluating mathematical reasoning of foundation models in visual contexts.
\newblock {\em arXiv preprint arXiv:2310.02255}, 2023.

\bibitem{zhang2024geoeval}
Jiaxin Zhang, Zhongzhi Li, Mingliang Zhang, Fei Yin, Chenglin Liu, and Yashar Moshfeghi.
\newblock Geoeval: benchmark for evaluating llms and multi-modal models on geometry problem-solving.
\newblock {\em arXiv preprint arXiv:2402.10104}, 2024.

\bibitem{wang2024measuring}
Ke~Wang, Junting Pan, Weikang Shi, Zimu Lu, Houxing Ren, Aojun Zhou, Mingjie Zhan, and Hongsheng Li.
\newblock Measuring multimodal mathematical reasoning with math-vision dataset.
\newblock {\em Advances in Neural Information Processing Systems}, 37:95095--95169, 2024.

\bibitem{zhang2024mathverse}
Renrui Zhang, Dongzhi Jiang, Yichi Zhang, Haokun Lin, Ziyu Guo, Pengshuo Qiu, Aojun Zhou, Pan Lu, Kai-Wei Chang, Yu~Qiao, et~al.
\newblock Mathverse: Does your multi-modal llm truly see the diagrams in visual math problems?
\newblock In {\em European Conference on Computer Vision}, pages 169--186. Springer, 2024.

\bibitem{li2024cmmath}
Zhong-Zhi Li, Ming-Liang Zhang, Fei Yin, Zhi-Long Ji, Jin-Feng Bai, Zhen-Ru Pan, Fan-Hu Zeng, Jian Xu, Jia-Xin Zhang, and Cheng-Lin Liu.
\newblock Cmmath: A chinese multi-modal math skill evaluation benchmark for foundation models.
\newblock {\em arXiv preprint arXiv:2407.12023}, 2024.

\bibitem{wang2025mv}
Peijie Wang, Zhong-Zhi Li, Fei Yin, Xin Yang, Dekang Ran, and Cheng-Lin Liu.
\newblock Mv-math: Evaluating multimodal math reasoning in multi-visual contexts.
\newblock {\em arXiv preprint arXiv:2502.20808}, 2025.

\bibitem{chatgpt}
OpenAI.
\newblock Chat{GPT}.
\newblock \url{https://openai.com/blog/chatgpt/}, 2022.

\bibitem{gemini}
Gemini Team, Rohan Anil, Sebastian Borgeaud, Yonghui Wu, Jean-Baptiste Alayrac, Jiahui Yu, Radu Soricut, Johan Schalkwyk, Andrew~M Dai, Anja Hauth, et~al.
\newblock Gemini: a family of highly capable multimodal models.
\newblock {\em arXiv preprint arXiv:2312.11805}, 2023.

\bibitem{claude3}
Anthropic.
\newblock Claude 3 models, 2024.
\newblock Accessed: 2025-05-06.

\bibitem{liu2023visual}
Haotian Liu, Chunyuan Li, Qingyang Wu, and Yong~Jae Lee.
\newblock Visual instruction tuning.
\newblock {\em Advances in neural information processing systems}, 36:34892--34916, 2023.

\bibitem{instructblip}
Wenliang Dai, Junnan Li, Dongxu Li, Anthony Meng~Huat Tiong, Junqi Zhao, Weisheng Wang, Boyang Li, Pascale Fung, and Steven Hoi.
\newblock Instructblip: Towards general-purpose vision-language models with instruction tuning.
\newblock {\em Advances in neural information processing systems}, 36:49250--49267, 2023.

\bibitem{bai2023qwen}
Jinze Bai, Shuai Bai, Shusheng Yang, Shijie Wang, Sinan Tan, Peng Wang, Junyang Lin, Chang Zhou, and Jingren Zhou.
\newblock Qwen-vl: A versatile vision-language model for understanding, localization, text reading, and beyond.
\newblock {\em arXiv preprint arXiv:2308.12966}, 1(2):3, 2023.

\bibitem{chen2024internvl}
Zhe Chen, Jiannan Wu, Wenhai Wang, Weijie Su, Guo Chen, Sen Xing, Muyan Zhong, Qinglong Zhang, Xizhou Zhu, Lewei Lu, et~al.
\newblock Internvl: Scaling up vision foundation models and aligning for generic visual-linguistic tasks.
\newblock In {\em Proceedings of the IEEE/CVF Conference on Computer Vision and Pattern Recognition}, pages 24185--24198, 2024.

\bibitem{ye2023mplugowl}
Qinghao Ye, Haiyang Xu, Guohai Xu, Jiabo Ye, Ming Yan, Yiyang Zhou, Junyang Wang, Anwen Hu, Pengcheng Shi, Yaya Shi, Chaoya Jiang, Chenliang Li, Yuanhong Xu, Hehong Chen, Junfeng Tian, Qian Qi, Ji~Zhang, and Fei Huang.
\newblock mplug-owl: Modularization empowers large language models with multimodality, 2023.

\bibitem{lu2024deepseek}
Haoyu Lu, Wen Liu, Bo~Zhang, Bingxuan Wang, Kai Dong, Bo~Liu, Jingxiang Sun, Tongzheng Ren, Zhuoshu Li, Hao Yang, et~al.
\newblock Deepseek-vl: towards real-world vision-language understanding.
\newblock {\em arXiv preprint arXiv:2403.05525}, 2024.

\bibitem{chen2024far}
Zhe Chen, Weiyun Wang, Hao Tian, Shenglong Ye, Zhangwei Gao, Erfei Cui, Wenwen Tong, Kongzhi Hu, Jiapeng Luo, Zheng Ma, et~al.
\newblock How far are we to gpt-4v? closing the gap to commercial multimodal models with open-source suites.
\newblock {\em Science China Information Sciences}, 67(12):220101, 2024.

\bibitem{textmonkey}
Yuliang Liu, Biao Yang, Qiang Liu, Zhang Li, Zhiyin Ma, Shuo Zhang, and Xiang Bai.
\newblock Textmonkey: An ocr-free large multimodal model for understanding document.
\newblock {\em arXiv preprint arXiv:2403.04473}, 2024.

\bibitem{dong2024internlmxcomposer24khdpioneeringlargevisionlanguage}
Xiaoyi Dong, Pan Zhang, Yuhang Zang, Yuhang Cao, Bin Wang, Linke Ouyang, Songyang Zhang, Haodong Duan, Wenwei Zhang, Yining Li, Hang Yan, Yang Gao, Zhe Chen, Xinyue Zhang, Wei Li, Jingwen Li, Wenhai Wang, Kai Chen, Conghui He, Xingcheng Zhang, Jifeng Dai, Yu~Qiao, Dahua Lin, and Jiaqi Wang.
\newblock Internlm-xcomposer2-4khd: A pioneering large vision-language model handling resolutions from 336 pixels to 4k hd, 2024.

\bibitem{wang2024qwen2vlenhancingvisionlanguagemodels}
Peng Wang, Shuai Bai, Sinan Tan, Shijie Wang, Zhihao Fan, Jinze Bai, Keqin Chen, Xuejing Liu, Jialin Wang, Wenbin Ge, Yang Fan, Kai Dang, Mengfei Du, Xuancheng Ren, Rui Men, Dayiheng Liu, Chang Zhou, Jingren Zhou, and Junyang Lin.
\newblock Qwen2-vl: Enhancing vision-language model's perception of the world at any resolution, 2024.

\bibitem{mplug-owl2}
Qinghao Ye, Haiyang Xu, Jiabo Ye, Ming Yan, Haowei Liu, Qi~Qian, Ji~Zhang, Fei Huang, and Jingren Zhou.
\newblock mplug-owl2: Revolutionizing multi-modal large language model with modality collaboration.
\newblock {\em arXiv preprint arXiv:2311.04257}, 2023.

\bibitem{xu2024llavacot}
Guowei Xu, Peng Jin, Hao Li, Yibing Song, Lichao Sun, and Li~Yuan.
\newblock Llava-cot: Let vision language models reason step-by-step, 2024.

\bibitem{zhao2024bba}
Xueliang Zhao, Xinting Huang, Tingchen Fu, Qintong Li, Shansan Gong, Lemao Liu, Wei Bi, and Lingpeng Kong.
\newblock Bba: Bi-modal behavioral alignment for reasoning with large vision-language models.
\newblock {\em arXiv preprint arXiv:2402.13577}, 2024.

\bibitem{deng2024r}
Linger Deng, Yuliang Liu, Bohan Li, Dongliang Luo, Liang Wu, Chengquan Zhang, Pengyuan Lyu, Ziyang Zhang, Gang Zhang, Errui Ding, et~al.
\newblock R-cot: Reverse chain-of-thought problem generation for geometric reasoning in large multimodal models.
\newblock {\em arXiv preprint arXiv:2410.17885}, 2024.

\bibitem{xu2025redstar}
Haotian Xu, Xing Wu, Weinong Wang, Zhongzhi Li, Da~Zheng, Boyuan Chen, Yi~Hu, Shijia Kang, Jiaming Ji, Yingying Zhang, et~al.
\newblock Redstar: Does scaling long-cot data unlock better slow-reasoning systems?
\newblock {\em arXiv preprint arXiv:2501.11284}, 2025.

\bibitem{team2025kimi}
Kimi Team, Angang Du, Bofei Gao, Bowei Xing, Changjiu Jiang, Cheng Chen, Cheng Li, Chenjun Xiao, Chenzhuang Du, Chonghua Liao, et~al.
\newblock Kimi k1. 5: Scaling reinforcement learning with llms.
\newblock {\em arXiv preprint arXiv:2501.12599}, 2025.

\bibitem{guo2025deepseek}
Daya Guo, Dejian Yang, Haowei Zhang, Junxiao Song, Ruoyu Zhang, Runxin Xu, Qihao Zhu, Shirong Ma, Peiyi Wang, Xiao Bi, et~al.
\newblock Deepseek-r1: Incentivizing reasoning capability in llms via reinforcement learning.
\newblock {\em arXiv preprint arXiv:2501.12948}, 2025.

\bibitem{peng2025lmm}
Yingzhe Peng, Gongrui Zhang, Miaosen Zhang, Zhiyuan You, Jie Liu, Qipeng Zhu, Kai Yang, Xingzhong Xu, Xin Geng, and Xu~Yang.
\newblock Lmm-r1: Empowering 3b lmms with strong reasoning abilities through two-stage rule-based rl.
\newblock {\em arXiv preprint arXiv:2503.07536}, 2025.

\bibitem{meng2025mm}
Fanqing Meng, Lingxiao Du, Zongkai Liu, Zhixiang Zhou, Quanfeng Lu, Daocheng Fu, Tiancheng Han, Botian Shi, Wenhai Wang, Junjun He, et~al.
\newblock Mm-eureka: Exploring the frontiers of multimodal reasoning with rule-based reinforcement learning.
\newblock {\em arXiv preprint arXiv:2503.07365}, 2025.

\bibitem{chen2025r1v}
Liang Chen, Lei Li, Haozhe Zhao, Yifan Song, and Vinci.
\newblock R1-v: Reinforcing super generalization ability in vision-language models with less than \$3.
\newblock \url{https://github.com/Deep-Agent/R1-V}, 2025.
\newblock Accessed: 2025-02-02.

\bibitem{liu2025visual}
Ziyu Liu, Zeyi Sun, Yuhang Zang, Xiaoyi Dong, Yuhang Cao, Haodong Duan, Dahua Lin, and Jiaqi Wang.
\newblock Visual-rft: Visual reinforcement fine-tuning.
\newblock {\em arXiv preprint arXiv:2503.01785}, 2025.

\bibitem{wang2025visualprm}
Weiyun Wang, Zhangwei Gao, Lianjie Chen, Zhe Chen, Jinguo Zhu, Xiangyu Zhao, Yangzhou Liu, Yue Cao, Shenglong Ye, Xizhou Zhu, et~al.
\newblock Visualprm: An effective process reward model for multimodal reasoning.
\newblock {\em arXiv preprint arXiv:2503.10291}, 2025.

\bibitem{liu2025othink}
Zhiyuan Liu, Yuting Zhang, Feng Liu, Changwang Zhang, Ying Sun, and Jun Wang.
\newblock Othink-mr1: Stimulating multimodal generalized reasoning capabilities through dynamic reinforcement learning.
\newblock {\em arXiv preprint arXiv:2503.16081}, 2025.

\bibitem{shen2025vlm}
Haozhan Shen, Peng Liu, Jingcheng Li, Chunxin Fang, Yibo Ma, Jiajia Liao, Qiaoli Shen, Zilun Zhang, Kangjia Zhao, Qianqian Zhang, et~al.
\newblock Vlm-r1: A stable and generalizable r1-style large vision-language model.
\newblock {\em arXiv preprint arXiv:2504.07615}, 2025.

\bibitem{malkinski2023review}
Miko{\l}aj Ma{\l}ki{\'n}ski and Jacek Ma{\'n}dziuk.
\newblock A review of emerging research directions in abstract visual reasoning.
\newblock {\em Information Fusion}, 91:713--736, 2023.

\bibitem{carpenter1990one}
Patricia~A Carpenter, Marcel~A Just, and Peter Shell.
\newblock What one intelligence test measures: a theoretical account of the processing in the raven progressive matrices test.
\newblock {\em Psychological review}, 97(3):404, 1990.

\bibitem{sternberg1977component}
Robert~J Sternberg.
\newblock Component processes in analogical reasoning.
\newblock {\em Psychological review}, 84(4):353, 1977.

\bibitem{logie2014visuo}
Robert~H Logie.
\newblock {\em Visuo-spatial working memory}.
\newblock Psychology Press, 2014.

\bibitem{hegarty1999types}
Mary Hegarty and Maria Kozhevnikov.
\newblock Types of visual--spatial representations and mathematical problem solving.
\newblock {\em Journal of educational psychology}, 91(4):684, 1999.

\bibitem{wu2024deepseekvl2mixtureofexpertsvisionlanguagemodels}
Zhiyu Wu, Xiaokang Chen, Zizheng Pan, Xingchao Liu, Wen Liu, Damai Dai, Huazuo Gao, Yiyang Ma, Chengyue Wu, Bingxuan Wang, Zhenda Xie, Yu~Wu, Kai Hu, Jiawei Wang, Yaofeng Sun, Yukun Li, Yishi Piao, Kang Guan, Aixin Liu, Xin Xie, Yuxiang You, Kai Dong, Xingkai Yu, Haowei Zhang, Liang Zhao, Yisong Wang, and Chong Ruan.
\newblock Deepseek-vl2: Mixture-of-experts vision-language models for advanced multimodal understanding, 2024.

\bibitem{Qwen2.5-VL}
Shuai Bai, Keqin Chen, Xuejing Liu, Jialin Wang, Wenbin Ge, Sibo Song, Kai Dang, Peng Wang, Shijie Wang, Jun Tang, Humen Zhong, Yuanzhi Zhu, Mingkun Yang, Zhaohai Li, Jianqiang Wan, Pengfei Wang, Wei Ding, Zheren Fu, Yiheng Xu, Jiabo Ye, Xi~Zhang, Tianbao Xie, Zesen Cheng, Hang Zhang, Zhibo Yang, Haiyang Xu, and Junyang Lin.
\newblock Qwen2.5-vl technical report.
\newblock {\em arXiv preprint arXiv:2502.13923}, 2025.

\bibitem{wang2024mpo}
Weiyun Wang, Zhe Chen, Wenhai Wang, Yue Cao, Yangzhou Liu, Zhangwei Gao, Jinguo Zhu, Xizhou Zhu, Lewei Lu, Yu~Qiao, and Jifeng Dai.
\newblock Enhancing the reasoning ability of multimodal large language models via mixed preference optimization.
\newblock {\em arXiv preprint arXiv:2411.10442}, 2024.

\bibitem{hurst2024gpt}
Aaron Hurst, Adam Lerer, Adam~P Goucher, Adam Perelman, Aditya Ramesh, Aidan Clark, AJ~Ostrow, Akila Welihinda, Alan Hayes, Alec Radford, et~al.
\newblock Gpt-4o system card.
\newblock {\em arXiv preprint arXiv:2410.21276}, 2024.

\bibitem{jaech2024openai}
Aaron Jaech, Adam Kalai, Adam Lerer, Adam Richardson, Ahmed El-Kishky, Aiden Low, Alec Helyar, Aleksander Madry, Alex Beutel, Alex Carney, et~al.
\newblock Openai o1 system card.
\newblock {\em arXiv preprint arXiv:2412.16720}, 2024.

\bibitem{anthropic2025claude37sonnet}
Anthropic.
\newblock Claude 3.7 sonnet system card.
\newblock \url{https://www.anthropic.com/claude-3-7-sonnet-system-card}, 2025.
\newblock Accessed: 2025-05-17.

\bibitem{deepmind2024gemini2flash}
{Google DeepMind}.
\newblock {Gemini 2.0 Flash Thinking}.
\newblock \url{https://deepmind.google/technologies/gemini/flash-thinking/}, 2024.
\newblock Accessed: 2025-05-17.

\bibitem{deepmind2025gemini2.5pro}
{Google DeepMind}.
\newblock {Gemini 2.5 Pro}.
\newblock \url{https://blog.google/technology/google-deepmind/gemini-model-thinking-updates-march-2025/}, 2025.
\newblock Accessed: 2025-05-17.

\bibitem{yu2024texthawk}
Ya-Qi Yu, Minghui Liao, Jihao Wu, Yongxin Liao, Xiaoyu Zheng, and Wei Zeng.
\newblock Texthawk: Exploring efficient fine-grained perception of multimodal large language models.
\newblock {\em arXiv preprint arXiv:2404.09204}, 2024.

\bibitem{zhang2024android}
Jiwen Zhang, Jihao Wu, Yihua Teng, Minghui Liao, Nuo Xu, Xiao Xiao, Zhongyu Wei, and Duyu Tang.
\newblock Android in the zoo: Chain-of-action-thought for gui agents.
\newblock {\em arXiv preprint arXiv:2403.02713}, 2024.

\bibitem{zhang2024ui}
Jiwen Zhang, Yaqi Yu, Minghui Liao, Wentao Li, Jihao Wu, and Zhongyu Wei.
\newblock Ui-hawk: Unleashing the screen stream understanding for gui agents.
\newblock {\em Preprints, manuscript/202408.2137}, 2024.

\bibitem{huang2025ocreasoning}
Mingxin Huang, Yongxin Shi, Dezhi Peng, Songxuan Lai, Zecheng Xie, and Lianwen Jin.
\newblock Ocr-reasoning benchmark: Unveiling the true capabilities of mllms in complex text-rich image reasoning.
\newblock {\em arXiv preprint arXiv:2505.17163}, 2025.

\bibitem{jiang2024marvel}
Yifan Jiang, Kexuan Sun, Zhivar Sourati, Kian Ahrabian, Kaixin Ma, Filip Ilievski, Jay Pujara, et~al.
\newblock Marvel: Multidimensional abstraction and reasoning through visual evaluation and learning.
\newblock {\em Advances in Neural Information Processing Systems}, 37:46567--46592, 2024.

\bibitem{xu2025visulogic}
Weiye Xu, Jiahao Wang, Weiyun Wang, Zhe Chen, Wengang Zhou, Aijun Yang, Lewei Lu, Houqiang Li, Xiaohua Wang, Xizhou Zhu, et~al.
\newblock Visulogic: A benchmark for evaluating visual reasoning in multi-modal large language models.
\newblock {\em arXiv preprint arXiv:2504.15279}, 2025.

\bibitem{song2025visualpuzzles}
Yueqi Song, Tianyue Ou, Yibo Kong, Zecheng Li, Graham Neubig, and Xiang Yue.
\newblock Visualpuzzles: Decoupling multimodal reasoning evaluation from domain knowledge.
\newblock {\em arXiv preprint arXiv:2504.10342}, 2025.

\bibitem{feng2025visualsphinx}
Yichen Feng, Zhangchen Xu, Fengqing Jiang, Yuetai Li, Bhaskar Ramasubramanian, Luyao Niu, Bill~Yuchen Lin, and Radha Poovendran.
\newblock Visualsphinx: Large-scale synthetic vision logic puzzles for rl.
\newblock {\em arXiv preprint arXiv:2505.23977}, 2025.

\bibitem{chia2024puzzlevqa}
Yew~Ken Chia, Vernon Toh~Yan Han, Deepanway Ghosal, Lidong Bing, and Soujanya Poria.
\newblock Puzzlevqa: Diagnosing multimodal reasoning skills of language models with abstract visual patterns.
\newblock {\em ACL}, 2024.

\bibitem{mitchell2311comparing}
M~Mitchell, AB~Palmarini, and A~Moskvichev.
\newblock Comparing humans, gpt-4, and gpt-4v on abstraction and reasoning tasks. arxiv 2023.
\newblock {\em arXiv preprint arXiv:2311.09247}.

\bibitem{tong2024eyes}
Shengbang Tong, Zhuang Liu, Yuexiang Zhai, Yi~Ma, Yann LeCun, and Saining Xie.
\newblock Eyes wide shut? exploring the visual shortcomings of multimodal llms.
\newblock In {\em Proceedings of the IEEE/CVF Conference on Computer Vision and Pattern Recognition}, pages 9568--9578, 2024.

\bibitem{yang2024mathglm}
Zhen Yang, Jinhao Chen, Zhengxiao Du, Wenmeng Yu, Weihan Wang, Wenyi Hong, Zhihuan Jiang, Bin Xu, and Jie Tang.
\newblock Mathglm-vision: solving mathematical problems with multi-modal large language model.
\newblock {\em arXiv preprint arXiv:2409.13729}, 2024.

\end{thebibliography}
}

\onecolumn
\appendix

\begin{center}
\large \textbf{APPENDIX}  
\end{center}

\section{Construction Details of VisuRiddles Benchmark}
\label{appedix:benchmark}
We construct the VisuRiddles benchmark through a structured pipeline comprising \textbf{Collection}, \textbf{Cleaning}, and \textbf{Consolidation}:

\textbf{Collection.} We manually collect a diverse set of visual reasoning problems from Chinese National Civil Service Examination, which form the five core categories of the benchmark. RAVEN and Sudoku samples are generated following the method\footnotemark[1]\footnotetext[1]{\href{https://github.com/WellyZhang/RAVEN}{https://github.com/WellyZhang/RAVEN}} and the protocols\footnotemark[2]\footnotetext[2]{\href{https://gitee.com/mxx11/sudoku}{https://gitee.com/mxx11/sudoku}}.

\textbf{Cleaning.} After initial collection, we recruited twelve trained annotators to verify and refine the dataset. This process involved identifying and removing duplicated questions, correcting answer key errors, and filtering out incomplete or noisy samples to ensure data consistency and reliability. Quality control included annotator training, multi-pass review, dispute resolution, and cross-validation on overlapping samples.

\textbf{Consolidation.} All cleaned samples were manually translated into English, covering question and answer. Each item was then categorized into its corresponding reasoning type based on expert explanations provided by the Fenbi educational platform. All data were standardized into a unified JSON format, supporting structured access to image content, metadata, and reasoning annotations.

\section{Details of VisuRiddles Synthesizer}
\subsection{Configuration of the VisuRiddles Synthesizer}

\begin{table*}[h]
  \caption{Configuration details of the VisuRiddles Synthesizer across seven reasoning categories.}
  \label{tab:appendix_table1}
  \centering
  {\small
  \begin{tabular}{lcccc}
    \toprule
    \cmidrule(r){1-2}
    Rule     &    Sub-rule  & Layout & Template & Icon   \\
    \toprule
    Numerical & Line, Curve, Angle, Cart, Space, Parts  & 1 & 1  & 5k   \\
    Stylistic &  AND, OR, XOR, XNOR  & 3 & 5  & 5k  \\
    Attribute & Element, Group  & 2 & 1  & 5k \\
    Positional & Translate, Rotate, Flip  & 4 & 7  & 5k \\
    \multirow{2}{*}{Spatial}   & Unfolding, Three-View, 3D-Reconstruction & \multirow{2}{*}{7} & \multirow{2}{*}{13}  & \multirow{2}{*}{5k} \\
                               &View-Consistency, Multiple-Views &   &    &     \\
    Sudoku    & Sudoku  & 1 & 5  & 9 \\
    RAVEN     & Raven  & 1 & 7  & 10k   \\
    \bottomrule
  \end{tabular}
  \label{type}
  }
\end{table*}

\subsection{Synthesis Pipeline of Riddles Instances}
\label{appendix:synthesis}
The synthesis pipeline for generating riddle instances with structured perceptual descriptions consists of two main stages: Instance Construction and API Labeling.

\textbf{Instance Construction}\\
Instance synthesis is responsible for generating the visual instances and corresponding perceptual descriptions. The detailed steps are as follows:

1. Rule Selection:\\
The synthesis process begins with rule selection. We first extract representative reasoning patterns from real-world standardized tests, such as civil service visual logic exams. These high-level rules are categorized into positional, stylistic, numerical,attribute,spatial,raven and sudoku types, and further decomposed into finer-grained sub-rules, each representing a distinct visual transformation or logical operation (e.g., ``simultaneous rotation and revolution'', or ``black-white XOR operation''). This decomposition enables modular and interpretable synthesis, allowing each instance to target a specific reasoning skill.

2. Element Configuration:\\
Once a sub-rule is selected, we proceed to define the visual components in a structured manner. This includes: (i) choosing a layout template (e.g., $3 \times 3$ matrix, progression row, etc.), (ii) selecting or auto-generating icon elements with specific visual properties (e.g., shape, color, direction), and (iii) assigning a background grid. Elements are configured in a rule-consistent way: for example, a “positional” sub-rule may require translation across rows, while a “stylistic” sub-rule may apply logical operations to shape fill. Each configuration ensures the pattern remains visually interpretable and solvable.

3. Rendering and Annotating:\\
Each question instance is constructed by applying parameterized transformations under defined constraints: a single correct answer, consistent application of one rule, and plausible distractors. The resulting sequence of visual instances is rendered into final image grids. In parallel, the framework generates structured \textbf{perceptual description} in JSON format, including layout metadata, element-level properties (e.g., object shape, position, state), and reasoning traces. These annotations bridge perception and reasoning, enabling models to explicitly learn visual abstraction rather than relying on black-box pattern matching.

\textbf{API Labeling}\\
Due to the poor performance of existing commercial models in abstract visual reasoning (AVR) tasks, especially when processing images, these models often suffer from significant hallucination issues and fail to generate correct reasoning annotations. Thus, directly feeding images into the model does not produce accurate results. To overcome this, we introduce a new step: first, we replace the images with corresponding perceptual descriptions and input these descriptions into a large model. This allows the model to generate CoT based on the perceptual descriptions. The detailed steps are as follows:

1. CoT Generation:\\
Perceptual descriptions are first extracted from the synthesized visual instances to supply fine-grained visual information. These descriptions are then provided to an advanced commercial model (Gemini 2.5-Flash-Think\footnotemark[3]\footnotetext[3]{\href{https://deepmind.google/models/gemini/flash/}{https://deepmind.google/models/gemini/flash/}} in this work), which generates a Chain-of-Thought (CoT) reasoning process and produces a corresponding answer.

2. Validation:\\
The generated answer is compared to the correct answer of the synthesized riddle. If the model’s output does not align with the correct answer, the data point is discarded, ensuring that only valid and accurate samples are included in the final dataset.

\subsection{Examples of Synthesized Riddles Instances}
\label{appendix:synthesis example}
\begin{figure}[H]
  \centering
  \caption{Examples of synthesized riddle instances in Numerical, Positional, and Attribute}
  \includegraphics[width=\linewidth]{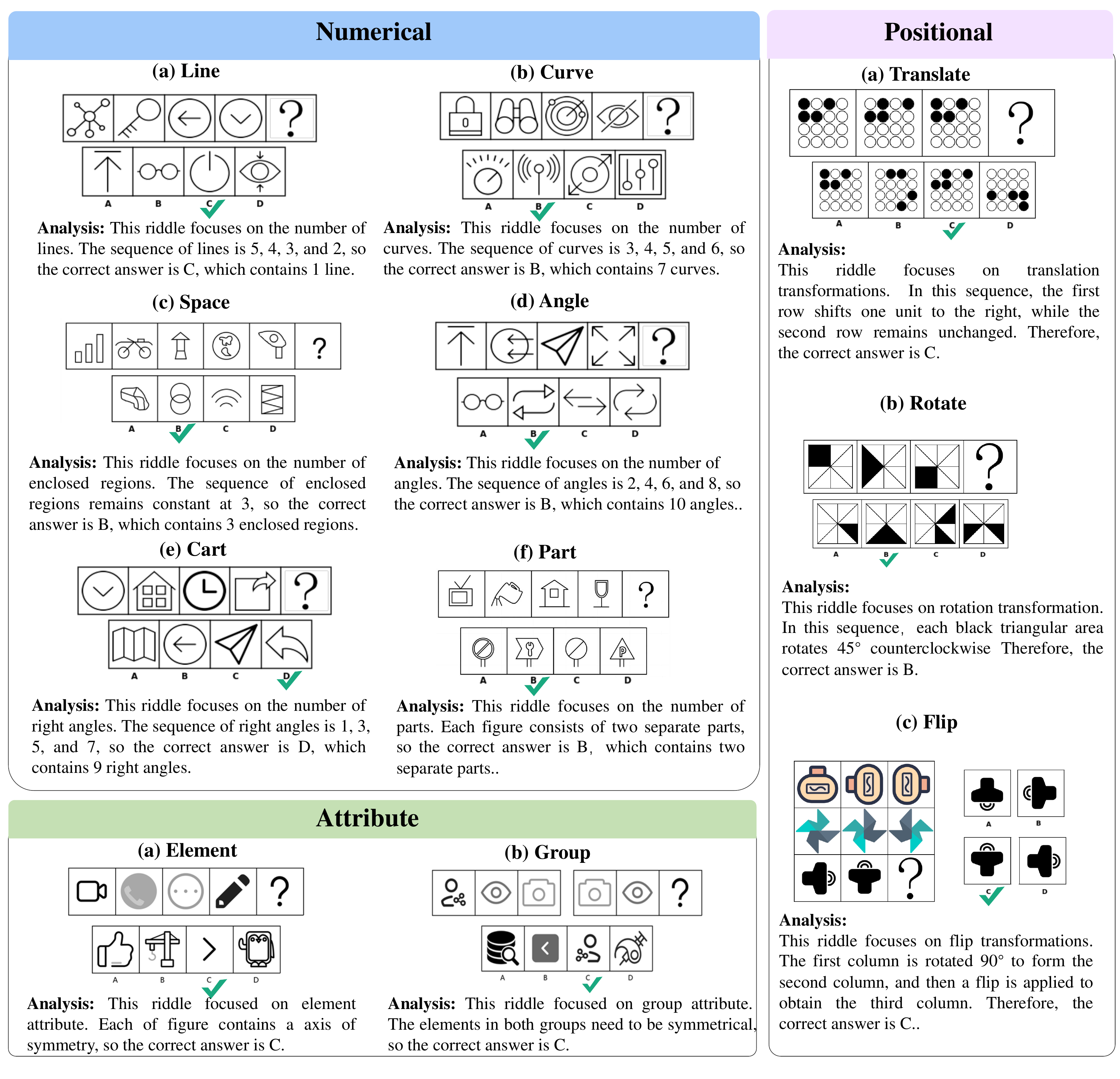}
  \label{fig:appendix_exp1}
\end{figure}

\begin{figure}[H]
  \centering
  \caption{Examples of synthesized riddle instances in Spatial, Stylistic, Sudoku, and RAVEN}
  \includegraphics[width=\linewidth]{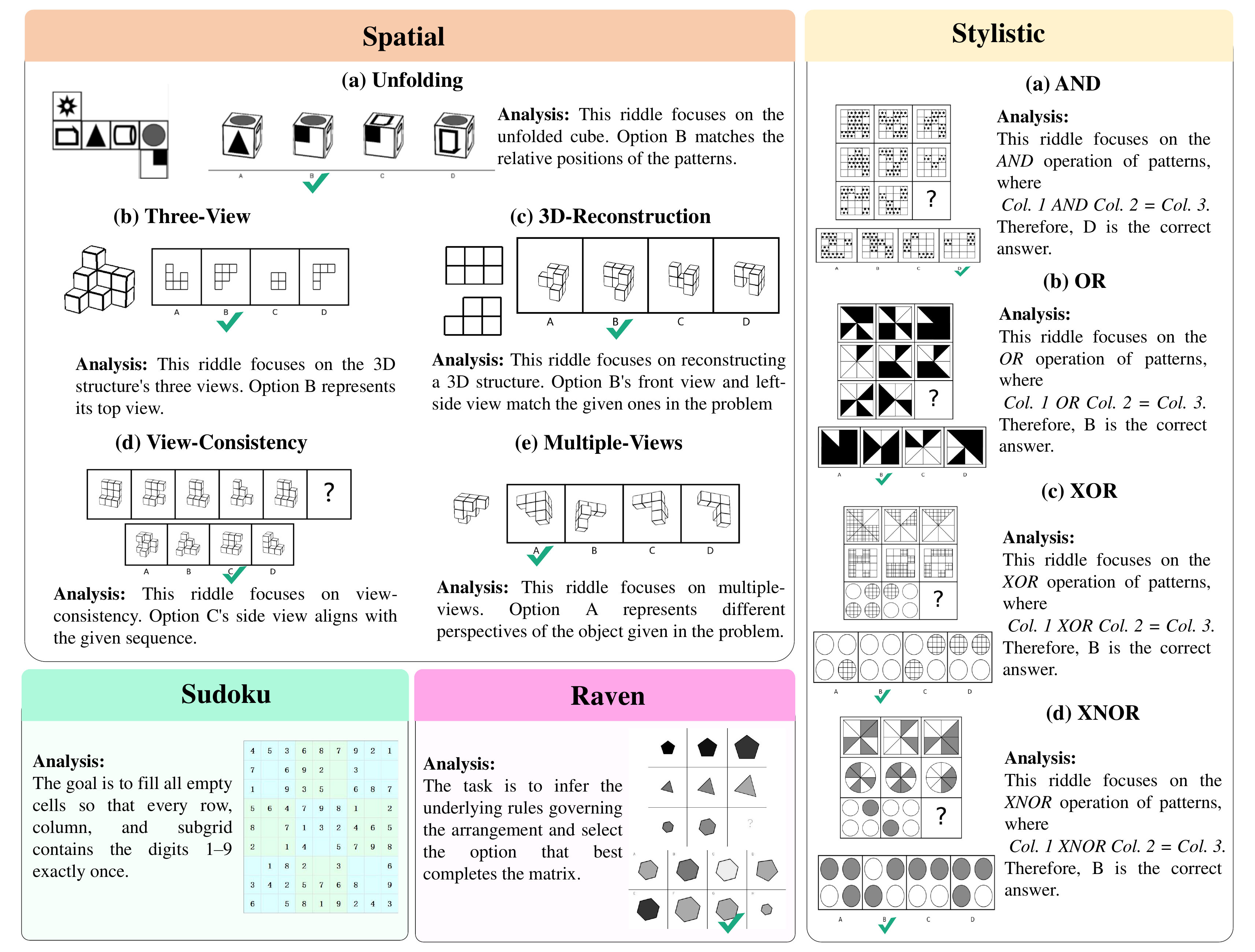}
  \label{fig:appendix_exp2}
\end{figure}

\newpage

\section{Details of Evaluated Models}
\label{evaluated_models}
We evaluate seventeen MLLMs, including eleven open-source models and six advanced proprietary models, as shown in Table ~\ref{appendix:mllms}. Human reference answers are obtained from the Fenbi Educational Platform, a widely used source for civil service exam preparation. Notably, GPT-o1, Claude-3-7-Sonnet-Thinking, and Gemini-2.5-Pro are equipped with specialized “thinking” modes aimed at enhancing multi-step reasoning. To further explore reasoning capabilities, CoT prompting is applied to GPT-4o, Qwen2.5-VL-72B, and InternVL2.5-78B-MPO, selected for their reasoning capacity, model scale, or lack of built-in “thinking” mode. Inference for all models is performed using the vLLM framework with unified decoding configurations.
\label{appendix:evaluation}
\begin{table}[h]
  \caption{Overview of evaluated MLLMs.}
  \label{appendix:mllms}
  \centering
  {\small
  \begin{tabular}{lccccc}
    \toprule
    \textbf{Model Name} & \textbf{Params} & \textbf{Think} & \textbf{CoT} & \textbf{Provider} & \textbf{Version} \\
    \midrule
    \multicolumn{6}{c}{\textit{Open-Source MLLMs}} \\
    \midrule
    Qwen2.5-VL-7B~\cite{Qwen2.5-VL}      & 7B   & ✗ & ✗ & Alibaba & - \\
    Qwen2.5-VL-32B      & 32B   & ✗ & ✗ & Alibaba & - \\
    Qwen2.5-VL-72B     & 72B  & ✗ & ✗ & Alibaba & - \\
    Qwen2.5-VL-72B(CoT)     & 72B  & ✗ & ✓ & Alibaba & - \\
    Qwen3-VL-235B-A22B-Instruct & 235B & ✗ & ✗ & Alibaba & - \\
    Qwen3-VL-235B-A22B-Thinking & 235B & ✓ & ✗ & Alibaba & - \\
    InternVL2.5-8B~\cite{chen2024expanding}      & 8B  & ✗ & ✗ & Shanghai AI Lab & - \\
    InternVL2.5-MPO-8B~\cite{wang2024mpo}      & 8B  & ✗ & ✗ & Shanghai AI Lab & - \\
    InternVL2.5-38B    & 38B  & ✗ & ✗ & Shanghai AI Lab & - \\
    InternVL2.5-38B-MPO  & 38B  & ✗ & ✗ & Shanghai AI Lab & - \\
    InternVL2.5-78B & 78B  & ✗ & ✗ & Shanghai AI Lab & - \\
    InternVL2.5-78B-MPO & 78B  & ✗ & ✗ & Shanghai AI Lab & - \\
    InternVL2.5-78B-MPO(CoT) & 78B  & ✗ & ✓ & Shanghai AI Lab & - \\
    DeepSeek-VL2~\cite{wu2024deepseekvl2mixtureofexpertsvisionlanguagemodels}& 32B  & ✗ & ✗ & DeepSeek & - \\
    MiniCPM-V 2.6~\cite{yao2024minicpm}      & 8B   & ✗ & ✗ & ModelBest & - \\
    Dots.vlm1 & 671B & ✗ & ✗ & Rednote hi lab & - \\
    \midrule
    \multicolumn{6}{c}{\textit{Commercial Products}} \\
    \midrule
    GPT-4o             & -  & ✗ & ✗ & OpenAI & 2024.11.20 \\
    GPT-4o(CoT)            & -  & ✗ & ✓ & OpenAI & 2024.11.20 \\
    o3            & -  & ✓ & ✗ & OpenAI & 2025.04.16 \\
    Claude-3.7-Sonnet  & -  & ✗ & ✗ & Anthropic & 2025.02.19 \\
    Claude-3.7-Sonnet-Thinking & - & ✓ & ✗ & Anthropic & 2025.02.19 \\
    Gemini-2.0-Flash-Thinking & - & ✓ & ✗ & Google DeepMind & 2025.01.21 \\
    Gemini-2.5-Flash-Thinking & - & ✓ & ✗ & Google DeepMind & 2025.06.21 \\
    Gemini-2.5-Pro     & -  & ✓ & ✗ & Google DeepMind & 2025.03.25 \\
    GPT-5 & -  & ✓ & ✗ & OpenAI & 2025.08.07 \\
    \bottomrule
  \end{tabular}
  }
\end{table}

\newpage

\section{Experimental result on VisuLogic}
\label{appendix:visulogic}
\begin{table}[H]
  \tiny
  \caption{\textbf{Experimental result on VisuLogic Benchmark.} The best results are marked \textbf{bold} and the second results are \underline{underlined}.}
 \label{sample-table-8x8}
  \centering
    \resizebox{\linewidth}{!}{
  \begin{tabular}{@{}l|c|cccccc|cc@{}}
    \toprule
    \textbf{Models} & \textbf{Param.}&    \textbf{Quantity} & \textbf{Spatiality} & \textbf{Position} & \textbf{Attribute} & \textbf{Style} & \textbf{Other} & \textbf{Overall} \\
    \midrule
    Human & -  & 45.3 & 52.7 & 71.1 & 50.0 & 47.5 & 44.2 &51.4 \\
    \midrule
    \multicolumn{8}{c}{\tiny{\textbf{MLLM Description$\rightarrow$LLM}}} \\
    \midrule
    Deepseek-R1 & 670B & 27.7 & 23.5 & 24.0 & 27.8 & 23.0 & \underline{35.0} & 26.6\\
    Qwen2.5-72B-Instruct &72B & 30.2 & 24.4 & 27.5 & 26.5 & 26.8 & 30.8 & 28.0\\
    Claude-3.7-Sonnet (20250219) &- & 26.6 & 22.5 & 25.0 & 28.0 & 25.6 & 30.6 & 25.9\\
    Doubao-1.5-Pro-32k (20250115) &- & 30.0 & 22.5 & 25.0 & 25.6 & 30.0 & 24.1 & 26.6\\
    \midrule
    \multicolumn{8}{c}{\textbf{Close Source MLLMs}} \\
    \midrule
    GPT-4o-mini (20240718) &- & 27.2 & 23.4 & 23.5 & 18.3 & 31.1 & 16.7 & 24.3\\
    GPT-4o (20240806) &- & 28.6 & 24.7 & 27.2 & 26.8 & 20.0 & 25.9 & 26.3\\
    Kimi-latest &- & 24.9 & 29.4 & 26.5 & 28.0& 16.7 & 26.9 & 25.9\\
    Doubao-1.5-Vision-Pro-32k (20250115)  &- & 28.1 & 23.8 & 29.1 & 25.1 & \underline{32.1} & \underline{35.0} & 28.1\\
    Gemini-2.0-Pro (20250205) &- & 29.7 & 24.2 & 27.9 & 30.5 & 22.2 & 33.3 &  28.0\\
    Claude-3.7-Sonnet (20250219) &- & 22.7 & 27.3 & 27.9 & 28.0 & 22.2 & 22.2 & 24.8\\
    \midrule
    \multicolumn{8}{c}{\tiny{\textbf{Open Source MLLMs}}} \\
    \midrule
    LLaVA-v1.5 &7B & 26.1 & 24.2 & 23.5 & 17.1 & 31.1 & 22.2 & 24.6\\
    LLaVA-OneVision (SI) & 7B  & 22.4 & 27.3 & 33.1 & 23.2 & 25.6 & 22.2 & 25.3\\
    Ovis2 &8B & 26.1 & 23.8 & 27.2 & 28.0 & 25.6 & 24.1 & 25.6\\
    Qwen2.5-VL-7B-Instruct &7B & 27.6 & 20.9 & 25.2 & 23.2 & \textbf{37.8} & 25.0 & 26.0\\
    Qwen2.5VL-72B-Instruct &72B & 25.2 & 23.8 & 27.2 & 25.6 & 25.6 & 34.3 & 26.2\\
    InternVL2.5-38B &8B & 24.4 & 26.4 & 27.2 & 23.2 & 25.6 & 26.9 & 25.5\\
    InternVL2.5-78B &78B & 26.6 & 26.0 & 26.5 & 26.8 & 31.1 & 30.6 & 27.3\\ 
    InternVL3-38B &38B & 28.7 & 27.6 & 26.1 & 21.4 & 23.9 & 28.5 & 27.1\\
    InternVL3-78B &78B  & 27.7 & 26.1 & \textbf{31.6} & 26.3 & 21.3 & 32.3 & 27.7\\
    Qwen2.5-VL-7B-Instruct-SFT (VisuLogic)  &7B & 24.4 & 26.4 & 27.2 & 23.2 & 25.6 & 26.9 & 25.5\\
    Qwen2.5-VL-7B-Instruct-RL (VisuLogic) &7B & 26.6 & \textbf{33.8} & \underline{29.4} & 23.2 & 18.9 & 29.6 & 28.0\\
    InternVL2.5-38B-SFT (VisuLogic) &38B & \underline{30.6} & 29.4 & 20.6 & 25.6 & 30.0 & 25.0 & 27.9\\
    InternVL2.5-38B-RL (VisuLogic)  &38B  & \textbf{31.2} & \underline{31.2} & 26.5 & \underline{30.5} & 30.0 & \textbf{38.9}& \textbf{31.1}\\
    \midrule
    \multicolumn{8}{c}{\tiny{\textbf{Ours}}} \\
    \midrule
    PAVR & 7B  & 28.9 & 30.3 & \textbf{31.6}  & \textbf{34.1} & 30.0 & 33.3 &\underline{30.6}\\
    \bottomrule
  \end{tabular}}
\end{table}

\end{document}